\documentclass[a4paper,12pt]{article}
\usepackage{lmodern}
\usepackage[T1]{fontenc}
\usepackage[utf8]{inputenc}
\usepackage[ngerman,english]{babel}
\usepackage{graphicx}
\graphicspath{{./esn-figs/}}
\usepackage[usenames]{xcolor}
\definecolor{fgreen}{rgb}{0.1,0.5,0.2}
\definecolor{grape}{rgb}{0.42,0.2,0.38}
\definecolor{marine}{rgb}{0,0.2,0.6}
\usepackage{tikz}
\usetikzlibrary{arrows.meta,shapes.geometric}
\tikzset{%
  >={Latex[width=2mm,length=2mm]},
          base/.style = {ellipse, draw=black,
                         minimum width=1.3cm, minimum height=0.8cm,
                         text centered, font=\sffamily},
       blister/.style = {circle, thick, draw=black,
                         minimum size=1.7cm, inner sep=0.1cm, outer sep=0.6,
                         text centered, font=\sffamily},
  reservoirSty/.style = {blister, fill=blue!20},
    readoutSty/.style = {base, fill=green!30},
}
\usepackage{enumitem}
\usepackage{dcolumn}
\usepackage{array}
\usepackage{caption}
\usepackage{lastpage}
\usepackage{amsmath}
\usepackage{amssymb}

\usepackage{listings}
\lstnewenvironment{lstcode}[1][]{\lstset{language=C++,#1}}{}
\usepackage{hyphenat}
\newcommand{\lsti}{\lstinline}
\PassOptionsToPackage{hyphens}{url}
\usepackage[bookmarks,bookmarksnumbered,colorlinks=true,linkcolor=grape,citecolor=fgreen,urlcolor=marine]{hyperref}
\newcommand{\cmpWidth}{0.97\textwidth}
\newcommand{\longTitle}{Modeling Biological Multifunctionality with Echo State Networks}  
\title{\bf\longTitle}
\author{Anastasia-Maria Leventi-Peetz$^1$, Jörg-Volker Peetz,\\
        Kai Weber, and Nikolaos Zacharis$^1$
  \\[1ex]
  {\footnotesize $^1$\,\parbox[t]{32em}{
    Department of Informatics \& Computer Engineering, Internet Computing and Cloud Technologies Lab, University of West Attica, ATH, GR}}
}
\date{{\small October, 2025}}
\hypersetup{
  pdfinfo={
    Author={A.-M. Leventi-Peetz, J.-V. Peetz, K. Weber, and N. Zacharis},
    Title={\longTitle}
  }
}
\begin{document}
\maketitle
\begin{abstract}
  \noindent
In this work, a three-dimensional multicomponent reaction-diffusion model has been developed, combining excitable-system dynamics with diffusion processes and sharing conceptual features with the \mbox{FitzHugh}-\mbox{Nagumo} model. Designed to capture the spatiotemporal behavior of biological systems, particularly electrophysiological processes, the model was solved numerically to generate time-series data. These data were subsequently used to train and evaluate an Echo State Network (ESN), which successfully reproduced the system’s dynamic behavior. The results demonstrate that simulating biological dynamics using data-driven, multifunctional ESN models is both feasible and effective.
\\*[1ex]
\end{abstract}

\section{Introduction}
Alan Turing's 1952 paper \emph{The Chemical Basis of Morphogenesis} is foundational work on reaction-diffusion (RD) models~\cite{b02}.
This paper demonstrates how local chemical interactions can give rise to global structure without centralized control, highlighting the emergence of self-organization, a core principle in complex systems. It provides a fundamental understanding of how organisms develop shape and structure through robust biological processes.
Applications of RD models are now well established across multiple domains. In biology and synthetic biology they are used to study tissue differentiation, tissue engineering, and the design of drug delivery systems. In materials science RD models contribute to the development of nanomaterials, smart materials, and active matter. In medical and disease modeling, they help simulate tumor growth, dynamic instabilities and feedback loops in neural circuits, processes that underlie both physiological and pathological activity patterns.
To address the limitations of RD models in biological contexts, where complexities often extend beyond simple activator-inhibitor interactions, hybrid approaches have been developed. These models integrate gene regulatory networks, positional information, cell signaling dynamics, and individual cell behaviors, including domain growth and boundary conditions. Such frameworks aim to more accurately reflect biological constraints and enable more sophisticated simulations. In its most general form, the RD model is expressed as a system of partial differential equations (PDEs) that describe how reactant concentrations evolve over space and time due to diffusion and chemical or biological interactions.
An emerging field of research applies artificial neural networks (ANNs) and in particular physics-informed neural networks (PINNs) to solve hybrid RD models that incorporate biological complexity, such as gene regulatory networks, cell signaling and tissue growth. PINNs embed physical laws directly into the loss function, making them especially well-suited for modeling complex RD systems~\cite{b01}.
In this direction, promising results have been published using frameworks like DeepXDE, SciANN, and NeuralPDE.jl~\cite{b07,b08,b09,b10}.
Extensions, such as biologically-informed neural networks (BINNs), a biologically grounded variant of PINNs, have also been introduced to model domain growth and uncover the underlying dynamics of biological systems from sparse experimental data~\cite{b03,b04,b05}.

\section{Reaction-Diffusion models and solutions}
Reaction-diffusion models are powerful tools for capturing the spatiotemporal dynamics of highly complex systems. Biological systems, in particular, exhibit multifunctional and nonlinear behavior that often challenges conventional analytical and numerical modeling approaches. In neuroscience, the multifunctionality of neural networks is reflected in their ability to switch between distinct dynamical states without altering synaptic properties, a phenomenon studied since the 1980s. More recent research has extended this concept to broader biological systems, aiming to uncover the mechanisms underlying such emergent behaviors.
Reservoir Computing (RC) represents a powerful paradigm for modeling multifunctionality in machine learning (ML), with Echo State Networks (ESNs) being particularly effective in emulating the temporal evolution and multifunctional behavior of biological processes. Their low training complexity, high memory capacity, and ability to generalize from limited data make ESNs well-suited for such tasks. A single ESN can simultaneously predict or simulate multiple interdependent variables, manage complex coupled dynamics, such as chemical, electrical, and genetic interactions, and support multi-output tasks within a unified framework.
Unlike traditional recurrent neural networks (RNNs), ESNs avoid the complexities of training recurrent connections by keeping the internal reservoir of interconnected neurons fixed. Only the connections between the reservoir and the output layer are trained, which significantly reduces computational cost and simplifies the learning process.
Thanks to their ability to retain memory of past inputs, a feature known as the \emph{echo state property}, ESNs can effectively process high-dimensional sequential data. This makes them particularly well-suited for time-series prediction and pattern recognition in complex biological datasets, such as neural recordings and other biosignals where patterns often depend on prior states.
Numerous studies have demonstrated the advantages of ESNs over traditional RNNs, particularly in applications that demand simplicity, computational efficiency, theoretical tractability, and biological plausibility.  ESNs have been successfully applied across a wide range of domains, from hydrology to EEG-based neuroscience, with recent innovations, such as DeepESNs, modular reservoir architectures and biologically inspired plasticity rules, further extending their capabilities and relevance.
Despite their demonstrated strengths, the application of ESNs in biology remains largely focused on neuroscientific and system-level dynamical modeling. Their use in molecular and physiological contexts, such as gene regulation, intracellular signaling, and metabolic dynamics, has been comparatively limited, representing an underexplored yet promising direction for future research~\cite{b12,b13,b14,b15,b18}.
ESNs have rarely been applied to the prediction of biological dynamics, such as modeling gene regulatory networks, signaling cascades, or protein concentration changes, processes that are central to developmental biology and regenerative medicine. Yet, ESNs are well-suited for modeling morphogenetic signaling and gene expression dynamics. Their intrinsic capacity to handle feedback loops, temporal delays, and hysteresis makes them particularly effective in capturing history-dependent behaviors, such as positional memory in cells. In this regard, ESNs offer distinct advantages for modeling the rich temporal structure of molecular and cellular processes, however, they are not inherently designed to incorporate physical laws or spatially structured inputs. As model-free systems by default, they lack built-in inductive biases that enforce physical constraints. Recent developments have begun to address these limitations and emerging approaches, such as physics-informed ESNs, spatially structured reservoir topologies, and hybrid models that integrate mechanistic components, offer promising avenues for extending ESNs to more physically grounded and biologically realistic applications~\cite{b11,b16,b18}.

\section{The electrophysiology model}
The FitzHugh-Nagumo (FHN) model is a widely used simplified excitable system for studying the emergence of spikes, traveling waves, and spatial patterns~\cite{b003,b004} . Extending the FHN model with the addition of diffusion terms, turns it into a reaction-diffusion system, capable of simulating excitations propagation across biological, chemical, and synthetic media.
Beyond its initial application in neuroscience, the FHN model now serves as a foundational tool for studying excitable phenomena across a wide range of scientific domains.
Applications include investigations of signal propagation and excitability thresholds in plant bioelectric signaling, wave dynamics during tissue morphogenesis, and the spatial regulation of protein activation at both cellular and tissue scales.
The RD model developed here shares key conceptual features with the FHN model, including nonlinear feedback, activator\hyp{}inhibitor dynamics, and the potential for oscillatory behavior. However, it extends the classical framework by incorporating a morphogen variable and sinusoidal coupling, resulting in a more general electrophysiological hybrid system.
Nonlinear RD electrophysiology hybrid models are increasingly relevant for modeling real biological systems, particularly in contexts where bioelectrical activity interacts with chemical signaling to drive complex behaviors such as pattern formation, morphogenesis, and excitability. Several practical biological applications employ, or propose, hybrid models of this kind or closely related conceptual frameworks.

\subsection{Model variables and equations}
Variables chosen for the model are: a chemical signal $c(x,y,z,t)$, an electrical potential $v(x,y,z,t)$, and an inhibitor $h(x,y,z,t)$. The chemical signal is represented by the concentration of a morphogen $c$, such as a signaling protein, an ion, or a diffusible metabolite. Electrical potentials can exist at both the cellular and tissue levels of non-neuronal systems. These include membrane potentials, serving as activators in model and bioelectric gradients that can influence cell behavior by modulating signaling pathways, directing cell migration, or triggering changes in gene expression. The inhibitor represents a molecule that suppresses or degrades the activator or signaling molecule, functioning as a diffusible antagonist, an enzyme that breaks down the signal, or a repressor protein within gene regulatory networks. A \emph{Reaction\hyp{}diffusion system} that describes the spatiotemporal evolution of the model variables can be written as:
\begin{align}
  \frac{\partial c}{\partial t} & = D_c \nabla^2 c + f(c, v, h)\nonumber\\
  \frac{\partial v}{\partial t} & = D_v \nabla^2 v + g(c, v, h) \label{eq:diffusion}\\
  \frac{\partial h}{\partial t} & = D_h \nabla^2 h + k(c, h)\nonumber
\end{align}
The functions $f$, $g$, and $k$ capture local biochemical interactions and depend on the model employed. Their choice ultimately determines whether the system exhibits oscillations, chaotic dynamics, or converges to a steady state.
Classical numerical methods such as Finite Differences (FD), Finite Elements (FE), and Finite Volumes (FV) are commonly used to solve electrophysiological model systems. As mentioned above, Artificial Intelligence (AI) and ML-techniques have also been employed for this purpose.
Equations~\ref{eq:diffusion} describe the dynamics of the interplay between electrical activity and physiological processes in an excitable system.
Many electrophysiological systems operate naturally near instability thresholds, where even subtle variations in timing or quantity can have significant effects. Accurately resolving their dynamics is essential for interpreting biological electrophysiology. ESNs are considered particularly well-suited for efficiently emulating transient system behavior.
The \textbf{electrophysiology model} with  the interaction functions in explicit form reads:
\begin{align}
   \frac{\partial c}{\partial t} & = D_c \nabla^2 c + \sigma v \sin c - \gamma c h \nonumber\\
   \frac{\partial v}{\partial t}  & = D_v \nabla^2 v +\sigma \sin (\chi c v) - \kappa h \label{eq:physmodel}\\
  \frac{\partial h}{\partial t} & =  D_h \nabla^2 h + \delta c - \eta h \nonumber
\end{align}
This nonlinear reaction-diffusion system captures:
\begin{itemize}
\item Nonlinear excitation through sine terms,
\item Inhibitory feedback, and
\item Spatiotemporal diffusion.
\end{itemize}
The nonlinear sine terms introduce rich oscillatory behavior characteristic of excitable systems like cardiac or neural tissue.
The bilinear inhibition term, $\gamma c h$ reflects competitive suppression, as seen in activator-inhibitor systems. The diffusion terms allow for spatial dynamics, wave propagation, and patterns creation. Also, $\chi$ is a scaling parameter applied to the nonlinear interaction between $c$ and $v$ modulating the frequency and amplitude of the nonlinear term. Increasing $\chi$ enhances the nonlinearity, potentially inducing stronger oscillations or chaotic transients, while smaller values yield a more linear response, possibly aiding numerical stability and convergence. The inhibitor $h$ follows a classical linear feedback model, activated by $c$ and decaying over time. The term $k(c,h) = \delta c - \eta h$ governs the dynamics of the inhibitor variable $h$, representing local reaction kinetics. It models the production of $h$ in response to the morphogen concentration $c$ and its natural decay, implementing a linear feedback mechanism typical of activator–inhibitor systems.
This model describes a system which is capable of producing \emph{Turing patterns}, oscillatory, and possibly also chaotic behavior, depending on the parameter choices, diffusion rates, and domain size.

\subsection{The model with ReservoirPy}
The Python library ReservoirPy~\cite{b23} is widely used to define, train and deploy RC architectures, such as ESNs. However, applying ReservoirPy to solve PDE systems in electrophysiology is not yet a common practice.
The typical workflow begins with spatial discretization of the PDEs which transforms the PDEs into a system of ordinary differential equations (ODEs) in time. Training data can be generated either by simulating the discretized model using traditional numerical solvers (e.g., those provided by SciPy) or by using existing datasets. An ESN can be trained to learn the system's complex time dynamics, serving as a surrogate model.
After training, the ESN can be used to generate time series predictions in an autoregressive fashion.
Taking $C^{t}$, $V^{t}$ and $H^{t}$ to represent the chemical, electrical, and inhibitory field states at time step $t$, and $\Delta t$ a temporal step size, the system's evolution can be calculated using the Euler method:
\paragraph{Discrete time updates:} (Euler method)
\begin{align*}
  C^{t+1} & = C^{t} + \Delta t \left[ D_c \nabla^2 C^{t} + f(C^{t}, V^{t}, H^{t}) \right]\\
  V^{t+1} & = V^{t} + \Delta t \left[ D_v \nabla^2 V^{t} + g(C^{t}, V^{t}, H^{t}) \right]\\
  H^{t+1} & = H^{t} + \Delta t \left[ D_h \nabla^2 H^{t} + k(C^{t}, H^{t}) \right]
\end{align*}
The \textbf{Laplacian} $\nabla^2$ is approximated using finite differences. A standard 3D stencil gives:
\begin{align*}
  \nabla^2 f(x,y,z) \approx \left[\right. & f(x - d, y, z) + f(x + d, y, z)\\
                                   & + f(x, y - d, z) + f(x, y + d, z)\\
                                   & + f(x, y, z - d) + f(x, y, z + d)\\
                                   & \left. - 6 f(x, y, z) \right] / d^3
\end{align*}
During training, the ESN receives the system’s state at time $t$ as input and is trained to predict the corresponding state at  $t+\Delta t$.
\paragraph{Echo State Network mapping:}
In ReservoirPy, connecting components (reservoirs and readouts) results in the creation of \emph{Model} objects. A model contains nodes in its components and maintains a description of all the connections between them. When executed, the model uses all the nodes and routes data according to the declared connections. During training, all readout layers (i.e., output layers within the model) are updated based on the incoming training data. In this work, a reservoir with 343 neurons was constructed to serve as the core of the ESN. An initial spatial vector $\boldsymbol{X}(0)$, representing the full state of the model system at time $0$ is considered. The ESN is fed a sequence of spatial vectors $\boldsymbol{X}(t)$ over time. The training target is to predict the next state, or a one-step-ahead forecast: $\boldsymbol{y}[t]=\boldsymbol{X}(t+\Delta t)$. This is a typical supervised learning setup in time-series modeling.
Linear regression, implemented in ReservoirPy via, e.g., a regularized ridge regression (called \emph{Ridge node}), is used to train the connections to the readout layer. This is an offline learning method, where the regression parameters are computed from all available input and target samples. 
The readout layer maps the reservoir state $\boldsymbol{x}[t]\in\mathbb{R}^{N}$ to the output $\boldsymbol{y}(t)\in\mathbb{R}^{M}$ with $\boldsymbol{W}_{\text{out}}$:
\begin{align*}
  \boldsymbol{W}_{\text{out}} = \boldsymbol{Y}\boldsymbol{X}^\intercal (\boldsymbol{X}\boldsymbol{X}^\intercal + \lambda \boldsymbol{I}_N)^{-1}
\end{align*}
Where $\boldsymbol{X}\in\mathbb{R}^{N\times T}$ is the matrix of reservoir activations (also referred to as reservoir states), $N$ the number of nodes in the reservoir and $T$ the number of training time steps. $\boldsymbol{I}_N$ stands for the unity matrix in $N\times N$.
The matrix of target values $\boldsymbol{Y}_{\text{target}}\in\mathbb{R}^{M\times T}$, contains the desired outputs at each time step, with  $M$ denoting the output dimensionality.

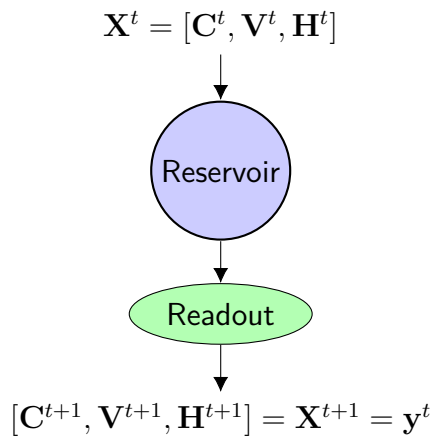
\begin{figure}[htbp]
  \centering
  \begin{tikzpicture}[node distance=1.3cm,
            every node/.style={font=\sffamily}, align=center]
    \node (Input)       [align=center]
                          {$\mathbf{X}^t = [\mathbf{C}^t, \mathbf{V}^t, \mathbf{H}^t]$};
    \node (Reservoir)   [reservoirSty,below of=Input,yshift=-0.6cm]
                          {Reservoir};
    \node (Readout)     [readoutSty,below of=Reservoir,yshift=-0.6cm]
                          {Readout};
    \node (Output)      [align=center,below of=Readout,yshift=-0.1cm]
                          {$[\mathbf{C}^{t+1}, \mathbf{V}^{t+1}, \mathbf{H}^{t+1}] = \mathbf{X}^{t+1} = \mathbf{y}^t$};
    \draw[->]         (Input) -- (Reservoir);
    \draw[->]         (Reservoir) -- (Readout);
    \draw[->]         (Readout) -- (Output);
    \end{tikzpicture}
  \caption{\label{f:esn-1}ESN architecture featuring shared input and output vectors and a single reservoir. The interdependence of the variables is captured by this reservoir, which is larger (contains more nodes) than each of the three reservoirs in the separate-input-channel architecture, depicted in Figure~\ref{f:esn-2}.}
\end{figure}

The parameter $\lambda\ > 0$ is a regularization coefficient that penalizes large output weights and helps prevent overfitting. The bias term, $\boldsymbol{b}$, if used, is stored in the node’s \lsti|Node.params| attribute. During inference, the Ridge node computes the output at time step $t$ using:
\begin{align*}
\boldsymbol{y}[t] = \boldsymbol{W}_{\text{out}} \cdot \boldsymbol{x}[t] + \boldsymbol{b}
\end{align*}
In the ESN, this operation corresponds to a linear readout, trained via linear regression to map reservoir states to output predictions.
An ESN can be created using various parameters, including the leaking rate of the reservoir neurons (\lsti|leak_rate|), the regularization coefficient for training the readout (\lsti|regularization_coef|), feedback connections from the outputs to the reservoir (\lsti|Wfb|), an activation function for the feedback (\lsti|fbfunc|), and a reference to a scikit-learn\footnote{A Python machine learning library.} linear regression model (\lsti|reg_model|). Different network structures with varying numbers of components, such as reservoirs, are also possible.
Figure~\ref{f:esn-1} illustrates a multifunctional ESN with a shared input and output vector and a single reservoir.
In Figure~\ref{f:esn-1}, the shared-input architecture with a single reservoir and shared output vector represents the configuration employed for all model versions and results presented in this work.

\paragraph{ESN setup:}
Inspiring for the ESN, was the multifunctional reservoir computing architecture, proposed by Mandal and Aihara~\cite{b06}.
Since the fields to be simulated are governed by coupled differential equations, their evolution is dynamically interdependent.
A multifunctional reservoir computing architecture should enable the simultaneous prediction of multiple fields by leveraging the intrinsic dependencies within the system. Different node configurations can be explored to identify the optimal setup for capturing the interdependencies among the input variables.

\begin{figure}[!htb]
\centering
\includegraphics[width=0.8\textwidth]{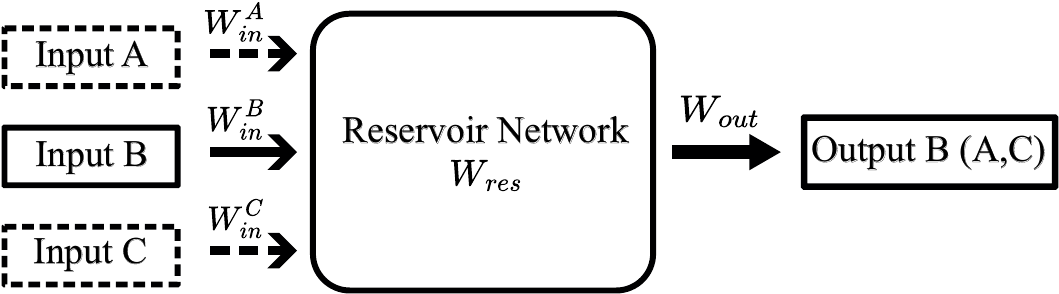}\\
\caption{\label{f:esn-0}Architecture proposed in Figure 1 of~\cite{b06}.}
\end{figure}

Mandal and Aihara~\cite{b06} proposed an architecture with separated input channels, each dedicated to a different network task, see Figure~\ref{f:esn-0}.
In~\cite{b06}, during the execution of a specific task, only the corresponding input channel is in this case active, while the others remained inactive. The reservoir and readout layers were designed to contain multifunctional neurons that participated in all tasks, whereas only the input layer included task-specific neurons.

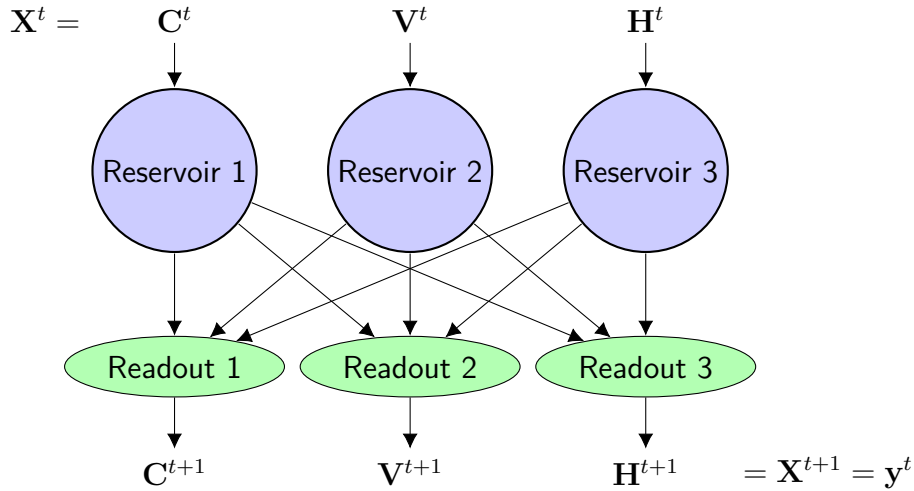
\begin{figure}[htbp]
  \centering
  \begin{tikzpicture}[node distance=1.3cm,
            every node/.style={font=\sffamily}, align=center]
    \node (Input1)      [align=center] {$\mathbf{C}^t$};
    \node (Input2)      [align=center, right of=Input1, xshift=1.8cm] {$\mathbf{V}^t$};
    \node (Input3)      [align=center, right of=Input2, xshift=1.8cm] {$\mathbf{H}^t$};
    \node (Reservoir1)  [reservoirSty,below of=Input1,yshift=-0.7cm] {Reservoir 1};
    \node (Reservoir2)  [reservoirSty,below of=Input2,yshift=-0.7cm] {Reservoir 2};
    \node (Reservoir3)  [reservoirSty,below of=Input3,yshift=-0.7cm] {Reservoir 3};
    \node (Readout1)    [readoutSty,below of=Reservoir1,yshift=-1.3cm] {Readout 1};
    \node (Readout2)    [readoutSty,below of=Reservoir2,yshift=-1.3cm] {Readout 2};
    \node (Readout3)    [readoutSty,below of=Reservoir3,yshift=-1.3cm] {Readout 3};
    \node (Output1)     [align=center,below of=Readout1,yshift=-0.1cm] {$\mathbf{C}^{t+1}$};
    \node (Output2)     [align=center,below of=Readout2,yshift=-0.1cm] {$\mathbf{V}^{t+1}$};
    \node (Output3)     [align=center,below of=Readout3,yshift=-0.1cm] {$\mathbf{H}^{t+1}$};
    \node (inputX)      [align=center, left of=Input1, xshift=-0.4cm] {$\mathbf{X}^t =$};
    \node (outputY)     [align=center,right of=Output3,xshift=1.1cm] {$= \mathbf{X}^{t+1} = \mathbf{y}^t$};
    \draw[->]         (Input1) -- (Reservoir1);
    \draw[->]         (Input2) -- (Reservoir2);
    \draw[->]         (Input3) -- (Reservoir3);
    \draw[->]         (Reservoir1) -- (Readout1);
    \draw[->]         (Reservoir1) -- (Readout2);
    \draw[->]         (Reservoir1) -- (Readout3);
    \draw[->]         (Reservoir2) -- (Readout1);
    \draw[->]         (Reservoir2) -- (Readout2);
    \draw[->]         (Reservoir2) -- (Readout3);
    \draw[->]         (Reservoir3) -- (Readout1);
    \draw[->]         (Reservoir3) -- (Readout2);
    \draw[->]         (Reservoir3) -- (Readout3);
    \draw[->]         (Readout1) -- (Output1);
    \draw[->]         (Readout2) -- (Output2);
    \draw[->]         (Readout3) -- (Output3);
    \end{tikzpicture}
  \caption{\label{f:esn-2}ESN architecture with separate input channels, three reservoirs, and three readouts. Each reservoir is connected to all readouts to model the interdependence among the inputs.}
\end{figure}

The separated-input-channel architecture was also tested in this work in the configuration shown in Figure~\ref{f:esn-2}.
Although this configuration produced reasonable yet less accurate predictions, the shared-input-channel architecture, shown in Figure~\ref{f:esn-1}, proved easier to optimize and yielded the best overall results.
For the shared input channel architecture, the three fields were merged into a single learning objective for the ESN, by concatenating all field components into a shared input vector for the reservoir.
At each time step $t$, the full system state is then represented by the (dicretized) spatial fields $C^t,V^t,H^t$, concatenated into a single input vector $\mathbf{X}^t = [C^t, V^t, H^t]$, fed into the ESN. The learning target is to predict the next system state $\mathbf{y}^t = \mathbf{X}^{t+1} = [C^{t+1}, V^{t+1}, H^{t+1}]$.
The dimension of this vector corresponds to the discretization of the 3D computational domain, given by $30\times 30\times 30= 27,000$ discretization points, multiplied by the number of field values per point which is $3$.
Training a unified ESN benefits from shared reservoir dynamics, which act as a regularizer to avoid overfitting. This setup allows the ESN to implicitly model interdependencies and temporal structure across all variables.
Three model versions have been calculated with slightly different parameters (see Table~\ref{tb:modParam}). The numerical results were used for training three ESN models with different parameters (see Table~\ref{tb:ESNparam}) and evaluating their predictive performance in regions of intensive system transformations and possibly chaos.
The input values for the ESN training are the exact numerical field values calculated during the first 100 simulation time steps for the models 1 and 2 and for the first 1000 time steps for model 3, respectively.
In Table~\ref{tb:ESNparam}, the parameter \lsti|WARM_UP|, also referred to as the \emph{washout period} or \emph{transient length}, specifies the number of initial time steps during which the reservoir is run but its states are not used for training. The purpose of this warm-up phase is to allow the reservoir to move beyond its initial transient dynamics, ensuring that training is based only on meaningful and representative internal states.
\begin{table}[htbp]
  \begin{minipage}[t]{\linewidth}
  \centering
  \setlength{\tabcolsep}{7pt}
  \begin{tabular}{lccc}
  \hline\hline
  Model &  version 1 &  version 2 & version 3 \\
  \hline
  $n_x\times n_y\times n_z$ & 30$\times$30$\times$30 &  30$\times$30$\times$30 &  30$\times$30$\times$30 \\
  $dx,dy,dz$      & 1.0 & 1.0 & 1.0 \\
  $dt$      & 0.01 & 0.01 & 0.002\\
  $D_c$     &  0.2 & 0.2 & 0.1 \\
  $D_v$     &  1.2 &  0.4 & 0.4 \\
  $D_h$     &  0.4 &  1.9 & 1.4 \\
  $\gamma$  & 2.5 &   2.5 & 2.5 \\
  $\delta$  & 2.0  &  2.0 & 2.0  \\
  $\eta$    & 1.5 &  1.5 & 1.5 \\
  $\kappa$  & 1.0 & 1.0 & 1.0 \\
  $\sigma$  & 2.0 & 2.0 & 2.0 \\
  $\chi$    &  1.0   & 9.1 & 8.5 \\
  \hline\hline
  \end{tabular}
  \end{minipage}
  \caption{\label{tb:modParam}Space discretization, step size, time step size and model parameters.}
\end{table}
\begin{table}[htbp]
  \begin{minipage}[t]{\linewidth}
  \centering
  \setlength{\tabcolsep}{7pt}
  \begin{tabular}{lccc}
  \hline\hline
  Model &  version 1 &  version 2 & version 3 \\
  \hline
  \lsti|UNITS|           &      343 &      343 & 343 \\
  \lsti|LEAK_RATE|       & $0.0002$ & $0.0002$ & $0.0001$ \\
  \lsti|SPECTRAL_RADIUS| &    $0.9$ &    $1.2$ & $1.4$ \\
  \lsti|WARM_UP|         &       50 &       45 & 275 \\ 
  \lsti|RIDGE|           & 1.e\,-7  & 1.e\,-11 & 1.e\,-11 \\
  trainings time slot    & $(0,100)$ & $(0,100)$ & $(0,1000)$ \\
  \hline\hline
  \end{tabular}
  \end{minipage}
  \caption{\label{tb:ESNparam}ESN parameters.}
\end{table}

\section{ESN calculations and results}
\begin{figure}[!htb]
\centering
\textsf{\footnotesize Numerical solution}\\
\includegraphics[width=0.97\textwidth, viewport=0 10 864 210, clip]{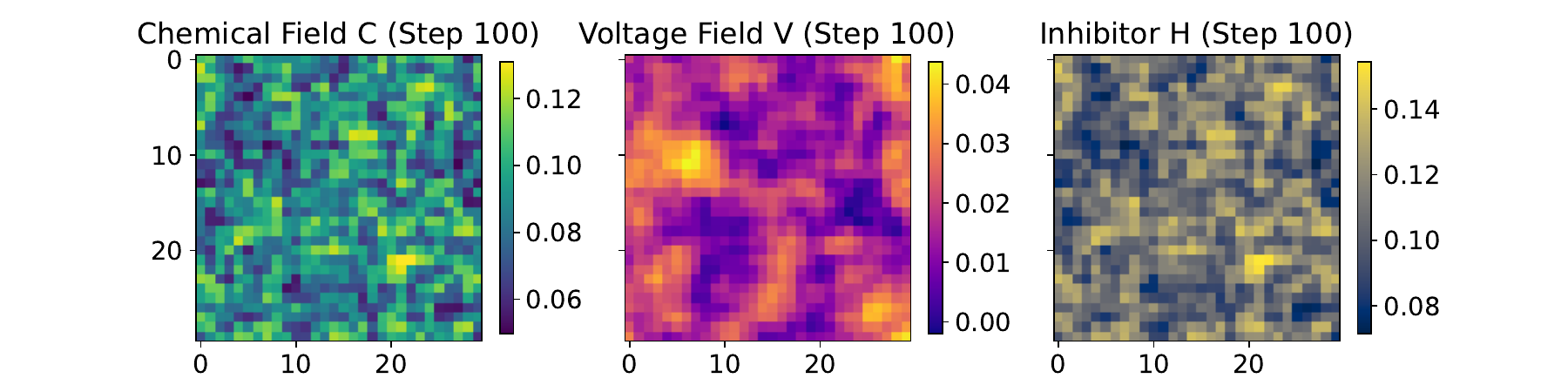}\\
\textsf{\footnotesize ESN prediction}\\
\includegraphics[width=0.97\textwidth, viewport=0 10 864 210, clip]{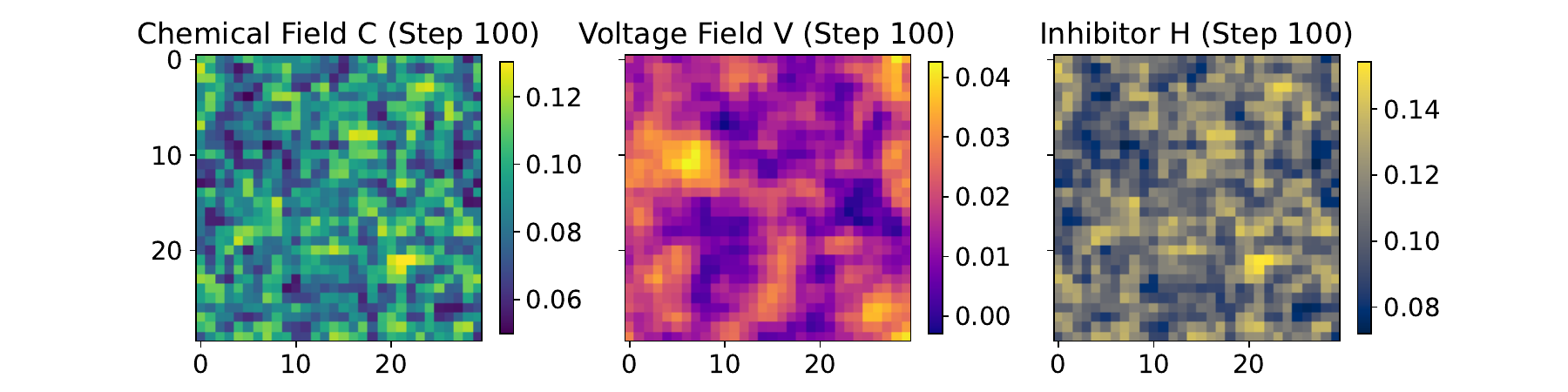}
\caption{\label{f:cmp02}Comparison of numerical simulation results and ESN predictions for time step 100, model version~1.}
\end{figure}

\begin{figure}[!htb]
\centering
\textsf{\footnotesize Numerical solution}\\
\includegraphics[width=0.97\textwidth, viewport=0 10 864 210, clip]{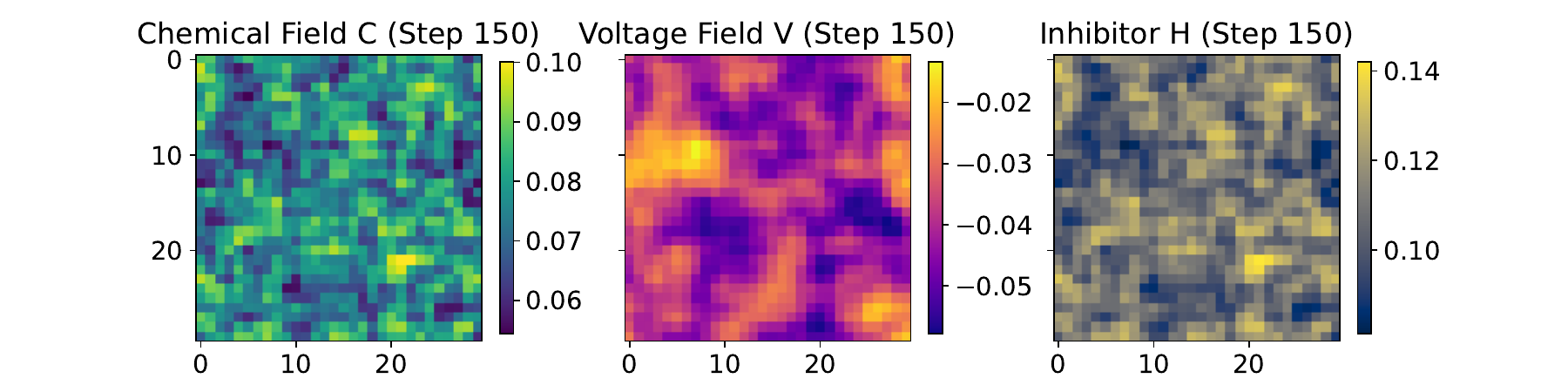}\\
\textsf{\footnotesize ESN prediction}\\
\includegraphics[width=0.97\textwidth, viewport=0 10 864 210, clip]{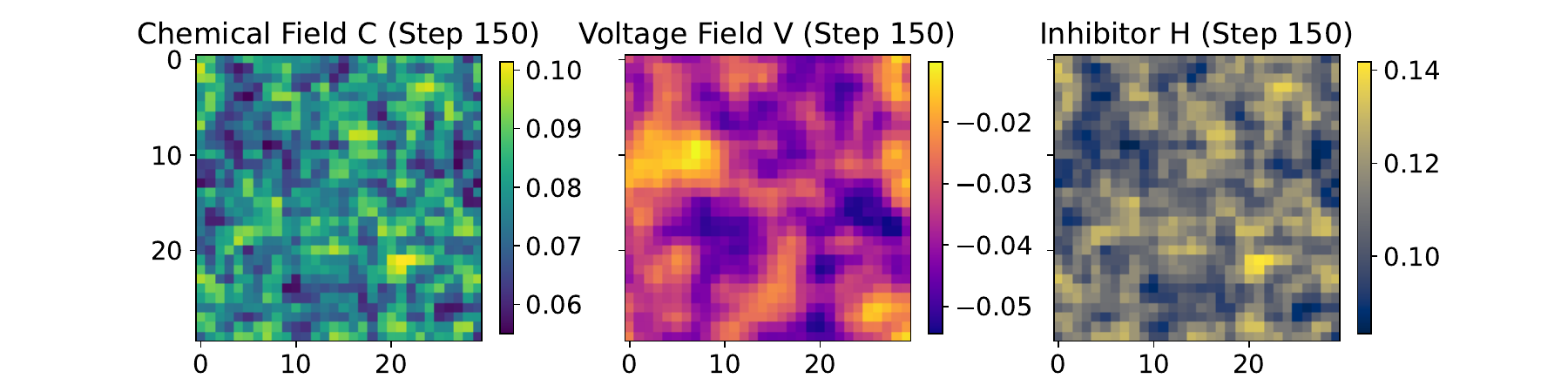}
\caption{\label{f:cmp03}Comparison of numerical simulation results and ESN predictions for time step 150, model version~1.}
\end{figure}

\begin{figure}[!htb]
\centering
\textsf{\footnotesize Numerical solution}\\
\includegraphics[width=\cmpWidth, viewport=0 10 864 210, clip]{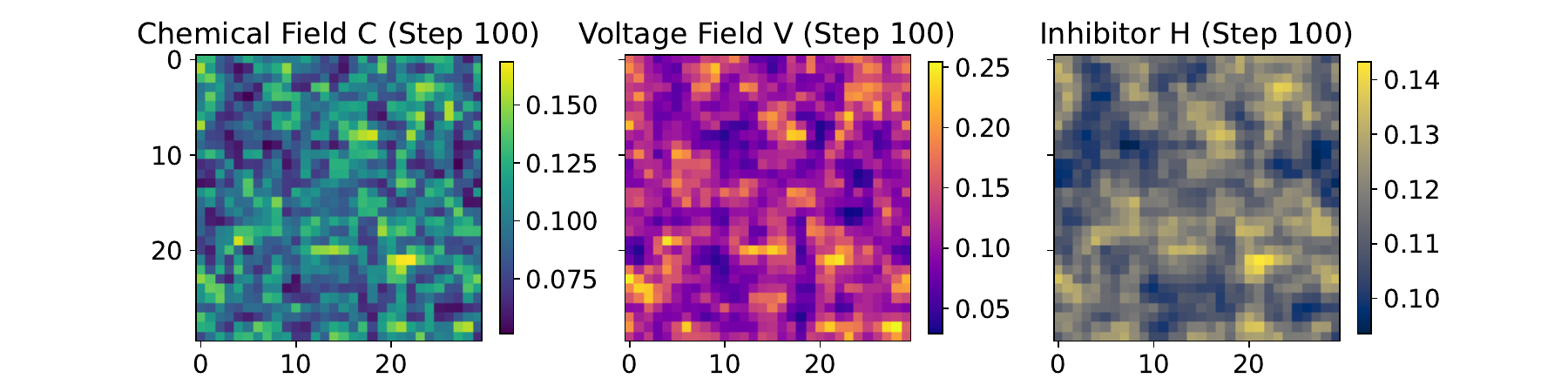}\\
\textsf{\footnotesize ESN prediction}\\
\includegraphics[width=\cmpWidth, viewport=0 10 864 210, clip]{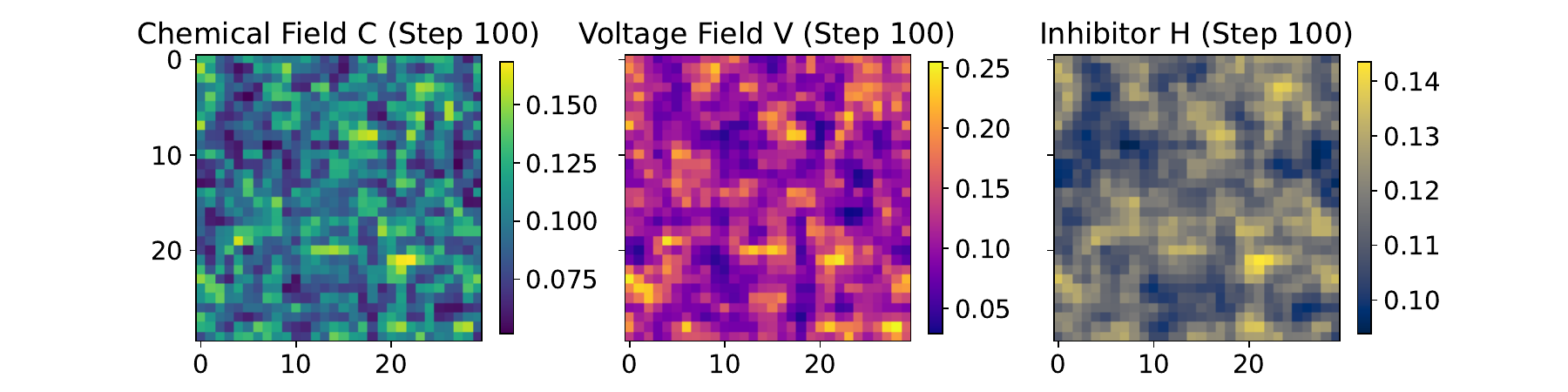}
\caption{\label{f:xcmp02}Comparison of numerical simulation results and ESN predictions for time step 100, model version~2.}
\end{figure}

\begin{figure}[!htb]
\centering
\textsf{\footnotesize Numerical solution}\\
\includegraphics[width=\cmpWidth, viewport=0 10 864 210, clip]{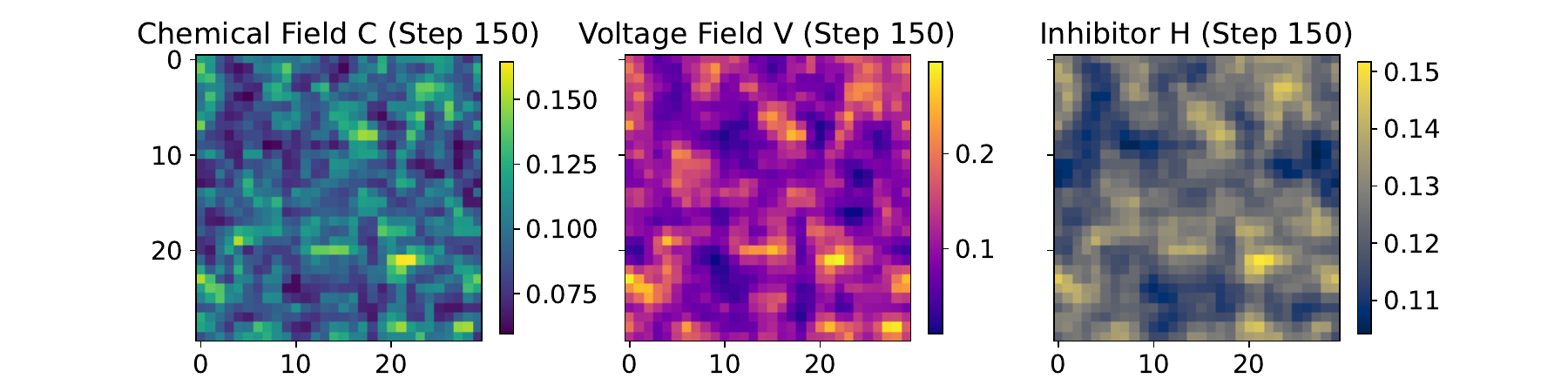}\\
\textsf{\footnotesize ESN prediction}\\
\includegraphics[width=\cmpWidth, viewport=0 10 864 210, clip]{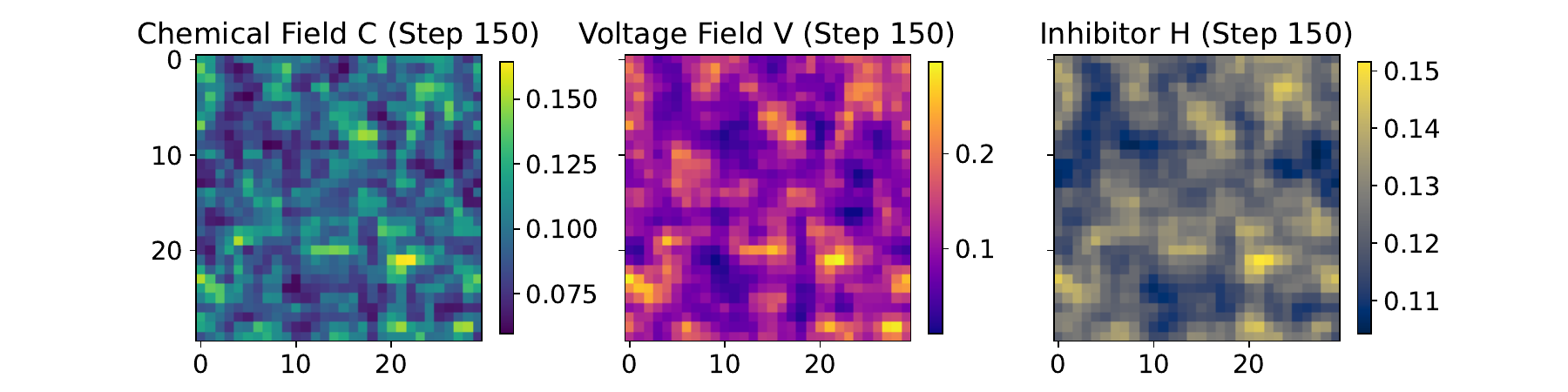}
\caption{\label{f:xcmp03}Comparison of numerical simulation results and ESN predictions for time step 150, model version~2.}
\end{figure}

\begin{figure}[!htb]
\centering
\textsf{\footnotesize Numerical solution}\\
\includegraphics[width=\cmpWidth, viewport=0 10 864 210, clip]{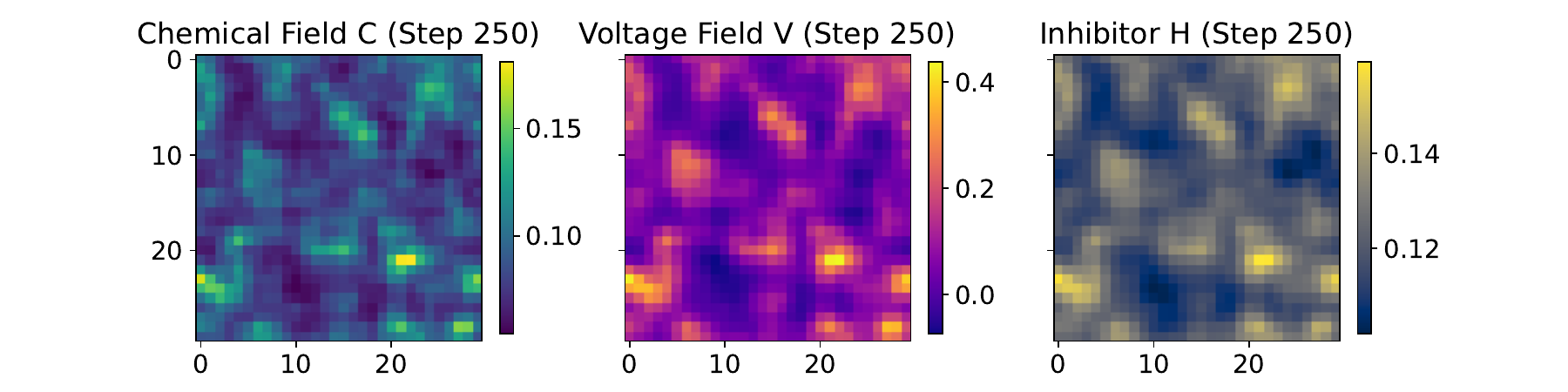}\\
\textsf{\footnotesize ESN prediction}\\
\includegraphics[width=\cmpWidth, viewport=0 10 864 210, clip]{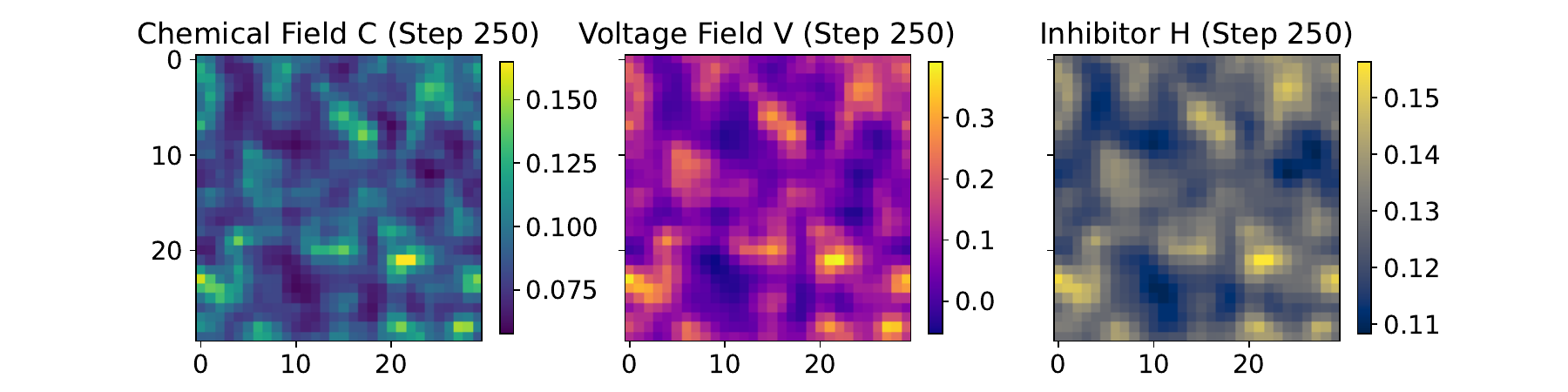}
\caption{\label{f:xcmp04}Comparison of numerical simulation results and ESN predictions for time step 250, model version 2.}
\end{figure}

\begin{figure}[!htb]
\centering
\textsf{\footnotesize Numerical solution}\\
\includegraphics[width=\cmpWidth, viewport=0 10 864 210, clip]{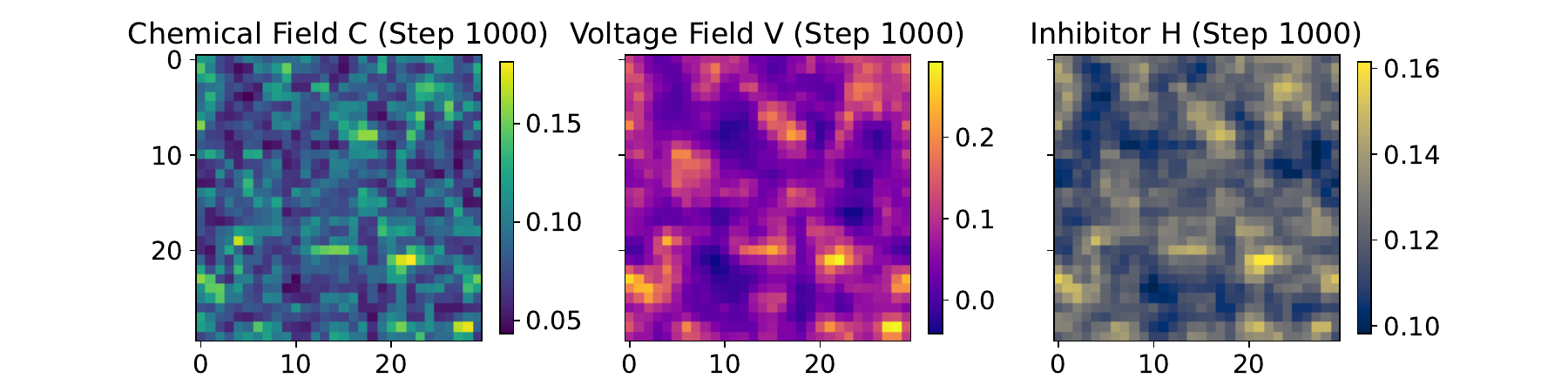}\\
\textsf{\footnotesize ESN prediction}\\
\includegraphics[width=\cmpWidth, viewport=0 10 864 210, clip]{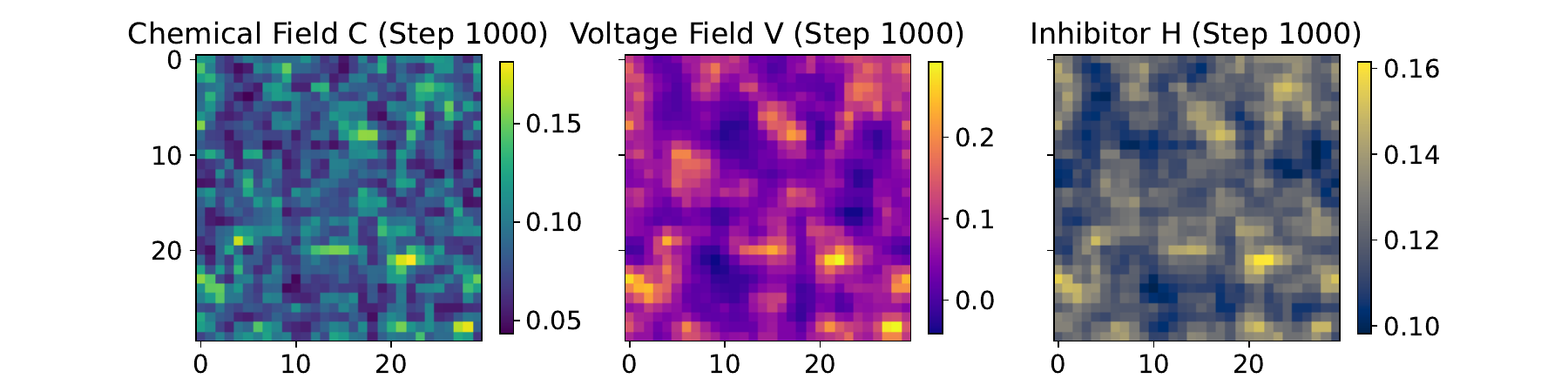}
\caption{\label{f:xcmp1005}Comparison of numerical simulation results and ESN predictions for time step 1000, model version~3.}
\end{figure}

\begin{figure}[!htb]
\centering
\textsf{\footnotesize Numerical solution}\\
\includegraphics[width=\cmpWidth, viewport=0 10 864 210, clip]{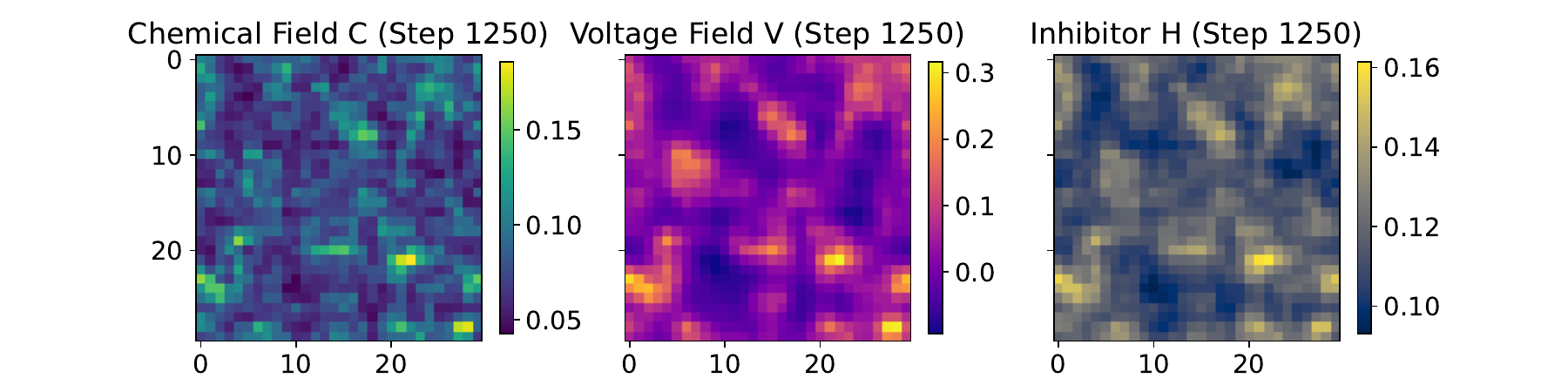}\\
\textsf{\footnotesize ESN prediction}\\
\includegraphics[width=\cmpWidth, viewport=0 10 864 210, clip]{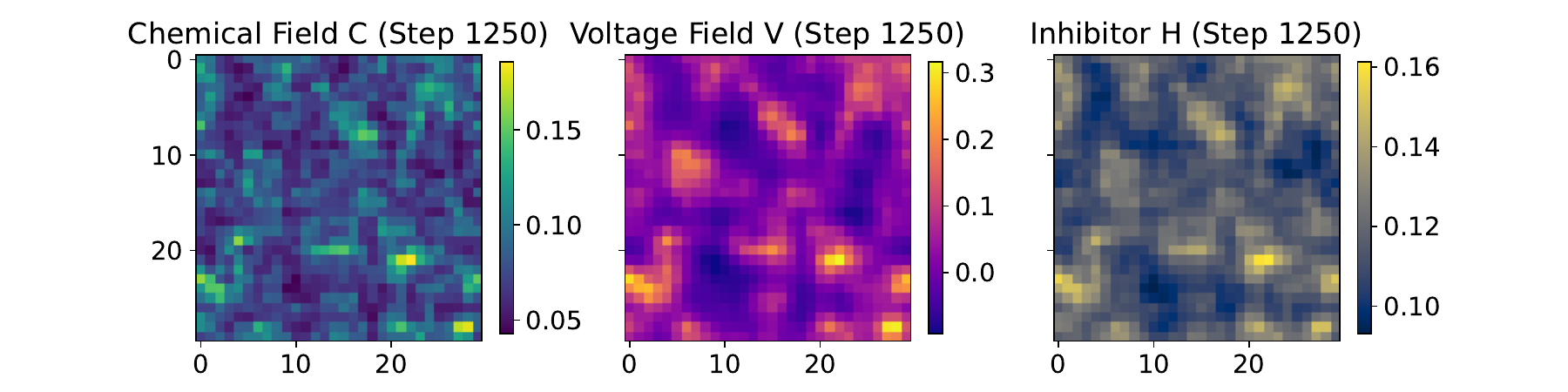}
\caption{\label{f:xcmp1205}Comparison of numerical simulation results and ESN predictions for time step 1250, model version~3.}
\end{figure}

\begin{figure}[!htb]
\centering
\textsf{\footnotesize Numerical solution}\\
\includegraphics[width=\cmpWidth, viewport=0 10 864 210, clip]{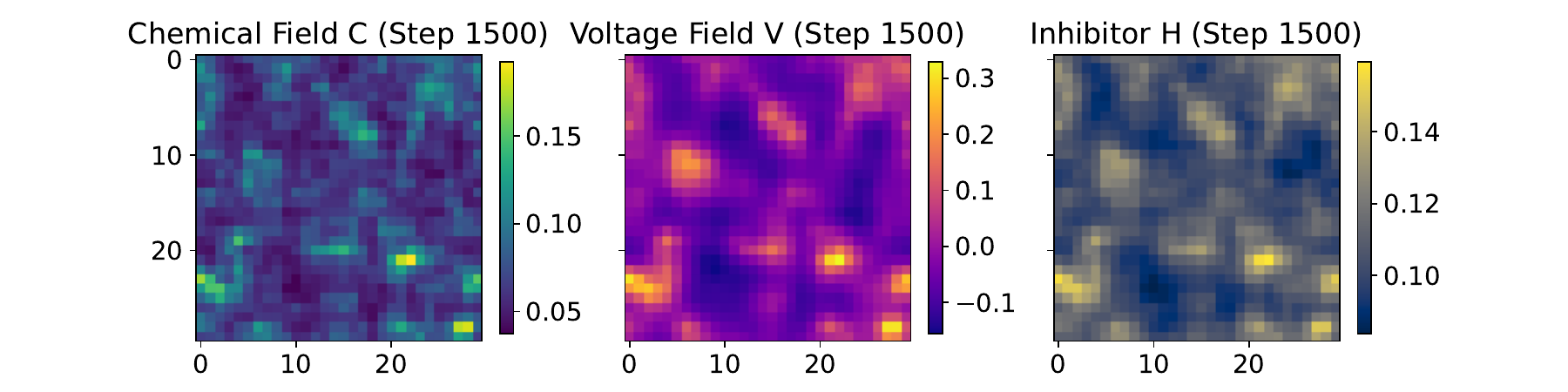}\\
\textsf{\footnotesize ESN prediction}\\
\includegraphics[width=\cmpWidth, viewport=0 10 864 210, clip]{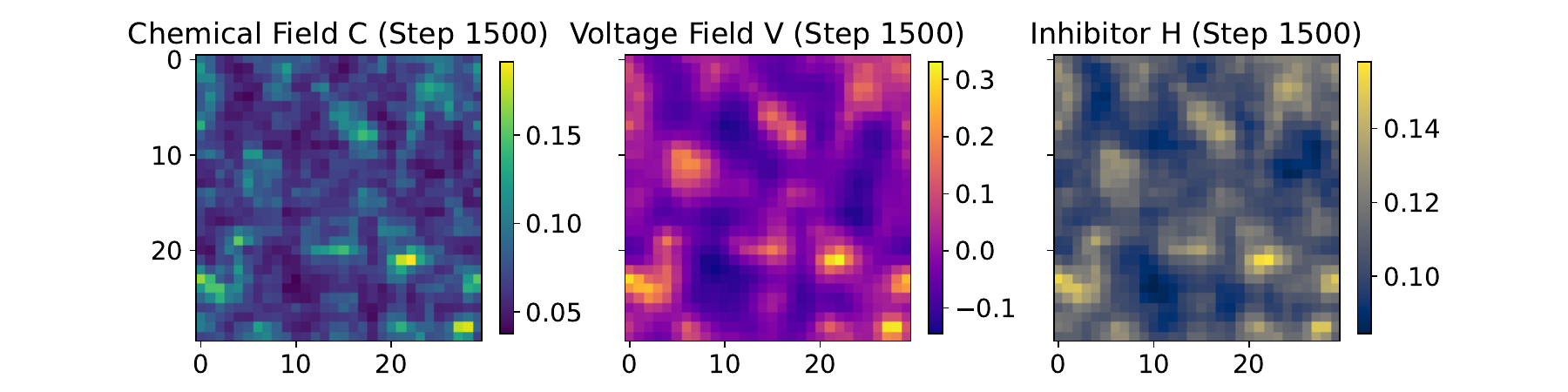}
\caption{\label{f:xcmp1505}Comparison of numerical simulation results and ESN predictions for time step 1500, model version~3.}
\end{figure}

To ensure consistency, comparisons between the numerically computed field values and the ESN predictions are presented as color maps on a fixed two-dimensional cross-sectional plane at $z=7$ across all visualizations. The color at each point indicates the value of the corresponding field according to its specific scale.
Comparisons at different time steps are displayed in Figures~\ref{f:cmp02} and \ref{f:cmp03} for model version~1; Figures~\ref{f:xcmp02}, \ref{f:xcmp03} and \ref{f:xcmp04} for model version~2; and Figures~\ref{f:xcmp1005},~\ref{f:xcmp1205} and \ref{f:xcmp1505} for model version~3. Each figure displays the distribution of field values over the fixed cross-sectional plane using color maps. For every model version, the two-dimensional color maps show the value distributions both during the training phase and at time steps following the end of training. The plots are densely filled due to the high spatial resolution of the simulation.
One can observe how coarse-grained structures gradually evolve into more distinct and organized patterns at later time steps. The early post-training phase is particularly noteworthy, as it reflects a period of intense transformation within the system. The predicted maps generated by the Echo State Network exhibit a high degree of similarity to the numerically computed reference maps, capturing both coarse and fine-scale structures with notable accuracy.
These results indicate that the ESN resolves the essential dynamics of the system and maintains predictive accuracy beyond the training period.

\begin{figure}[!htb]
\centering
\textsf{\footnotesize Lyapunov Diagnostics}\\
\includegraphics[width=0.9\textwidth, viewport=0 14 720 278, clip]{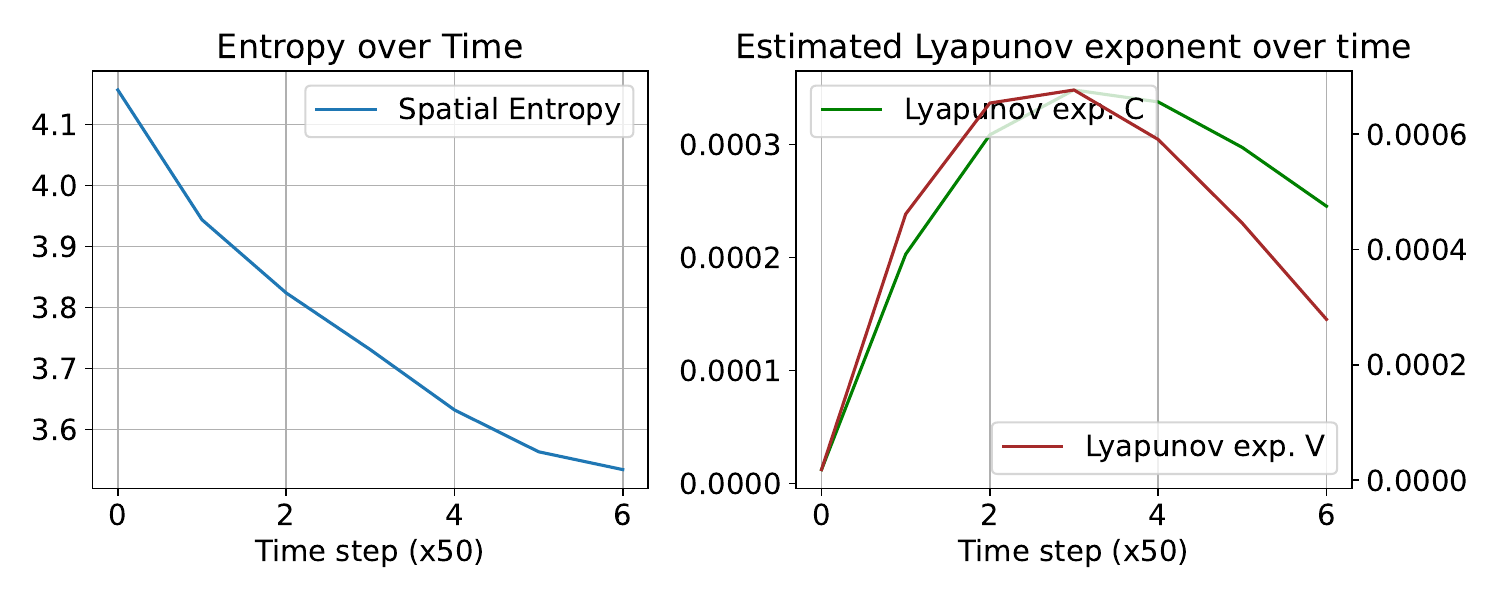}
\caption{\label{f:lyapD}Lyapunov exponent and entropy over time for the model version~1.}
\end{figure}

\begin{figure}[!htb]
\centering
\textsf{\footnotesize Lyapunov Diagnostics}\\
\includegraphics[width=0.9\textwidth, viewport=0 14 720 278, clip]{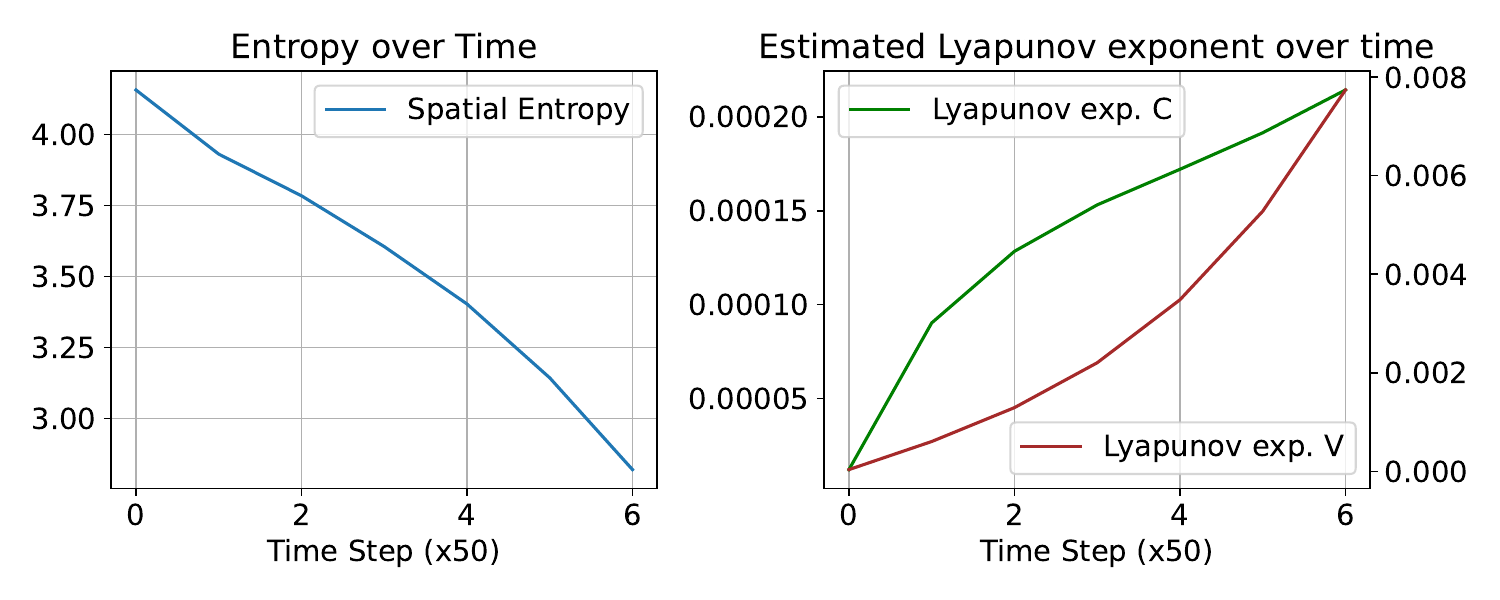}
\caption{\label{f:xlyapD}Lyapunov exponent and entropy over time for the model version~2.}
\end{figure}

\begin{figure}[!htb]
\centering
\textsf{\footnotesize Lyapunov Diagnostics}\\
\includegraphics[width=0.9\textwidth, viewport=0 14 720 278, clip]{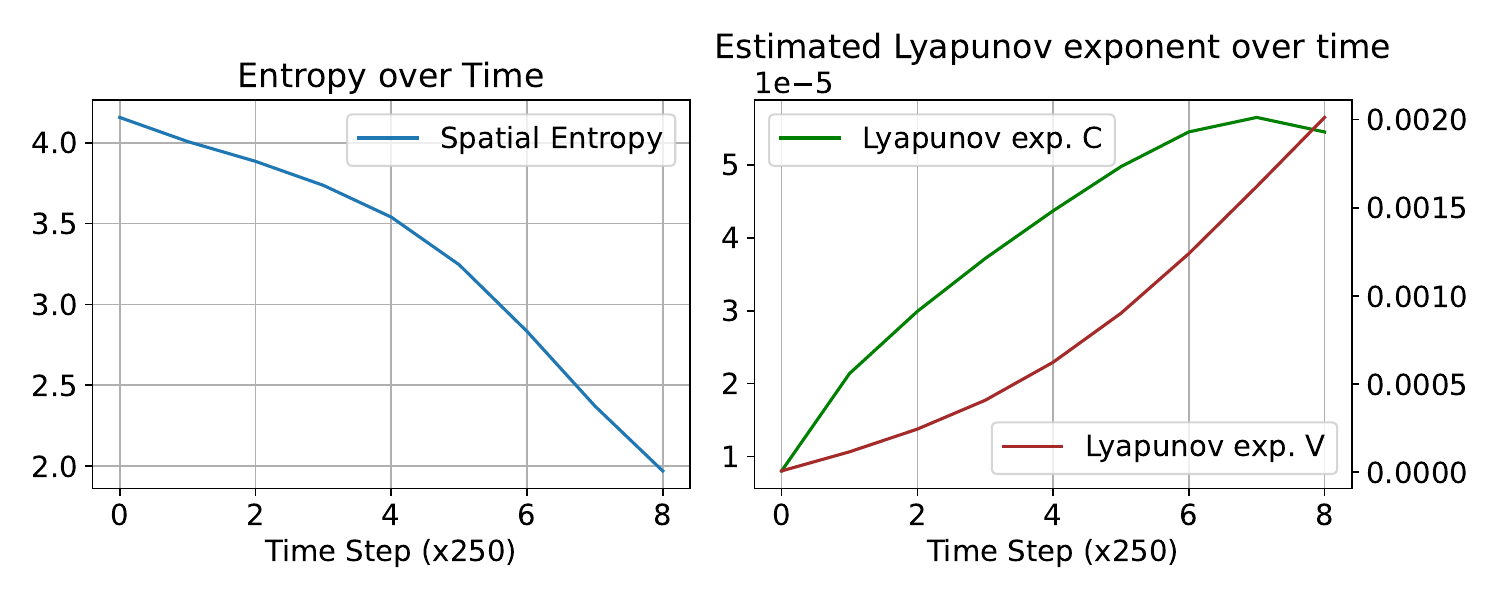}
\caption{\label{f:xlyapD05}Lyapunov exponent and entropy over time for model version~3.}
\end{figure}

The Lyapunov diagnostics and spatial entropy over time for the three model versions are shown in Figure~\ref{f:lyapD},~\ref{f:xlyapD} and~\ref{f:xlyapD05}, respectively.

Spatial entropy, computed from the distribution of field values across the computational domain, reflects the degree of the system's spatial complexity.
Decreasing entropy indicates a system gradually becoming more spatially ordered, while high or fluctuating entropy reflects persistent spatial complexity.

\begin{figure}[!htb]
\centering
\includegraphics[width=0.85\textwidth, viewport=0 16 864 850, clip]{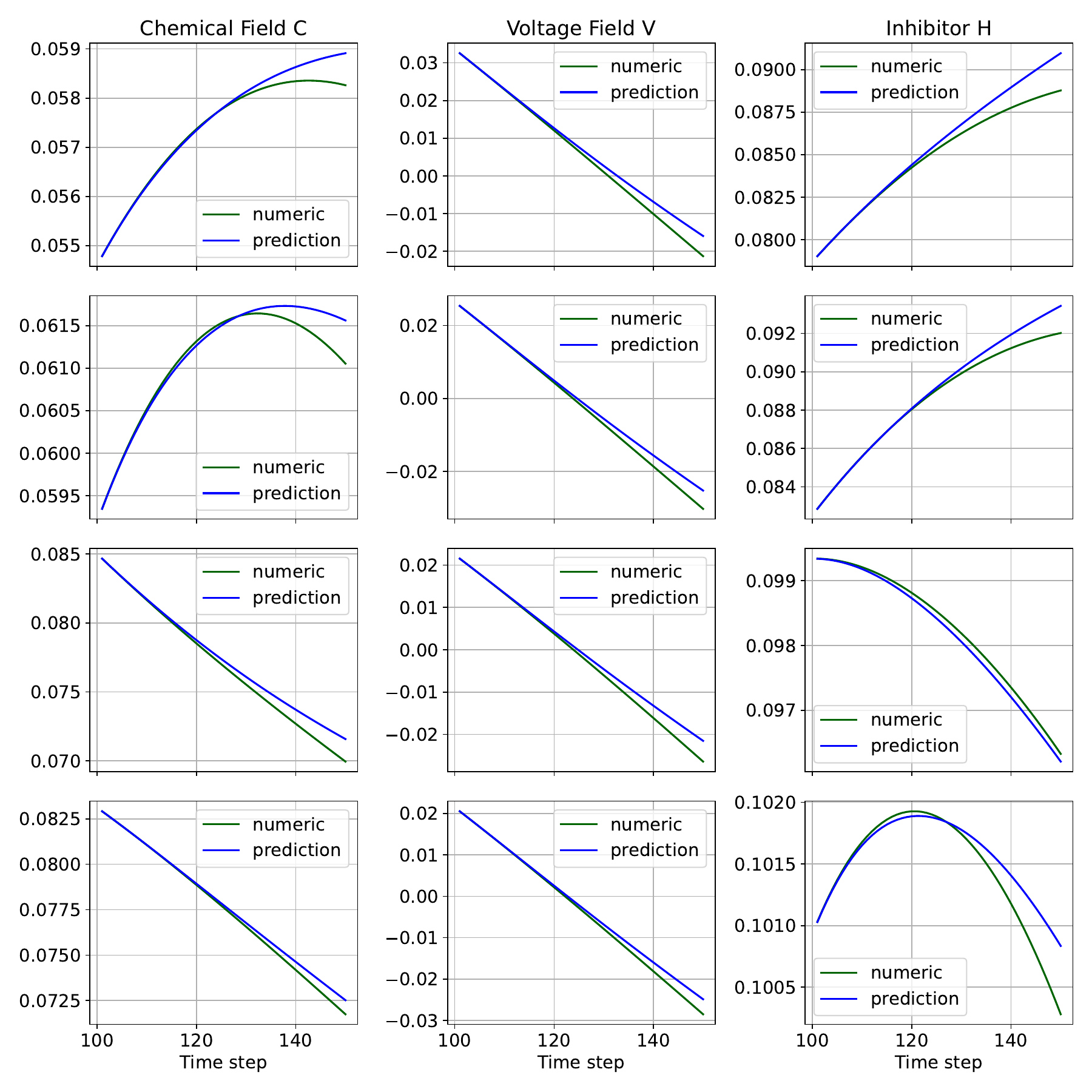}
\caption{\label{f:timecurves}Numerical simulation versus ESN predictions from step 100 to step 150 and for 4 points on the cross-section $z=7$ for model version~1.}
\end{figure}

\newcolumntype{x}{>{\centering\arraybackslash}p{0.17\textwidth}}
\begin{table}[htbp]
  \begin{minipage}[t]{\linewidth}
  \centering
  \setlength{\tabcolsep}{7pt}
  \begin{tabular}{rrr}
  \hline\hline
  model 1 & model 2 & model 3 \\
  \hline
  {[ 9, 4, $z$]} & {[ 9, 4, $z$]} & {[ 9, 4, $z$]} \\
  {[ 3,11, $z$]} & {[ 8, 8, $z$]} & {[ 3,11, $z$]} \\
  {[ 8, 8, $z$]} & {[ 9, 9, $z$]} & {[ 9, 9, $z$]} \\
  {[ 9, 9, $z$]} & {[14,13, $z$]} & {[14,13, $z$]} \\
                  & {[ 7,14, $z$]} & {[ 7,14, $z$]} \\
                  & {[ 7,20, $z$]} & {[ 7,20, $z$]} \\
  \hline\hline
  \end{tabular}
  \end{minipage}
  \caption{\label{tb:comp-points}Arbitrarily selected grid points on the visualization plane {$z$=7} for comparisons between ESN predictions and numerical field calculations.}
\end{table}
Grid points on the plane $z=7$ were arbitrarily selected for comparing ESN field predictions with numerical results across all three model versions (see Table~\ref{tb:comp-points}).
Figure~\ref{f:timecurves} shows comparisons between the numerical results of model version~1 and the corresponding ESN-predicted field values over a series of time steps, evaluated at the points selected for model~1 (see Table~\ref{tb:comp-points}).

\begin{figure}[!htb]
\centering
\includegraphics[width=0.85\textwidth, viewport=0 16 864 1280, clip]{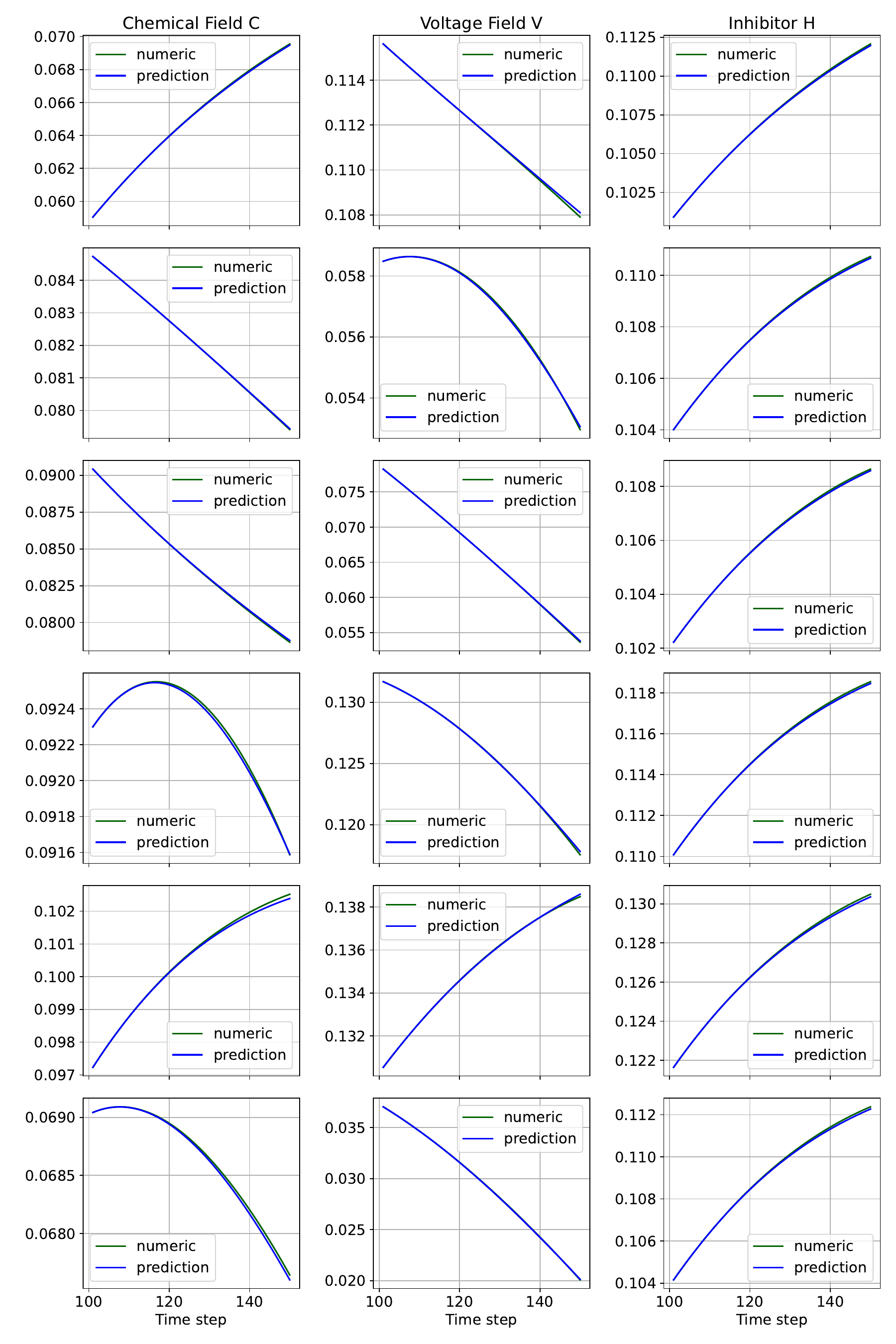}
\caption{\label{f:xtimecurves}Numerical simulation versus ESN predictions from step 100 to step 150 and for 6 points on the cross-section $z=7$ for model version~2.}
\end{figure}

\begin{figure}[!htb]
\centering
\includegraphics[width=0.85\textwidth, viewport=0 16 864 1280, clip]{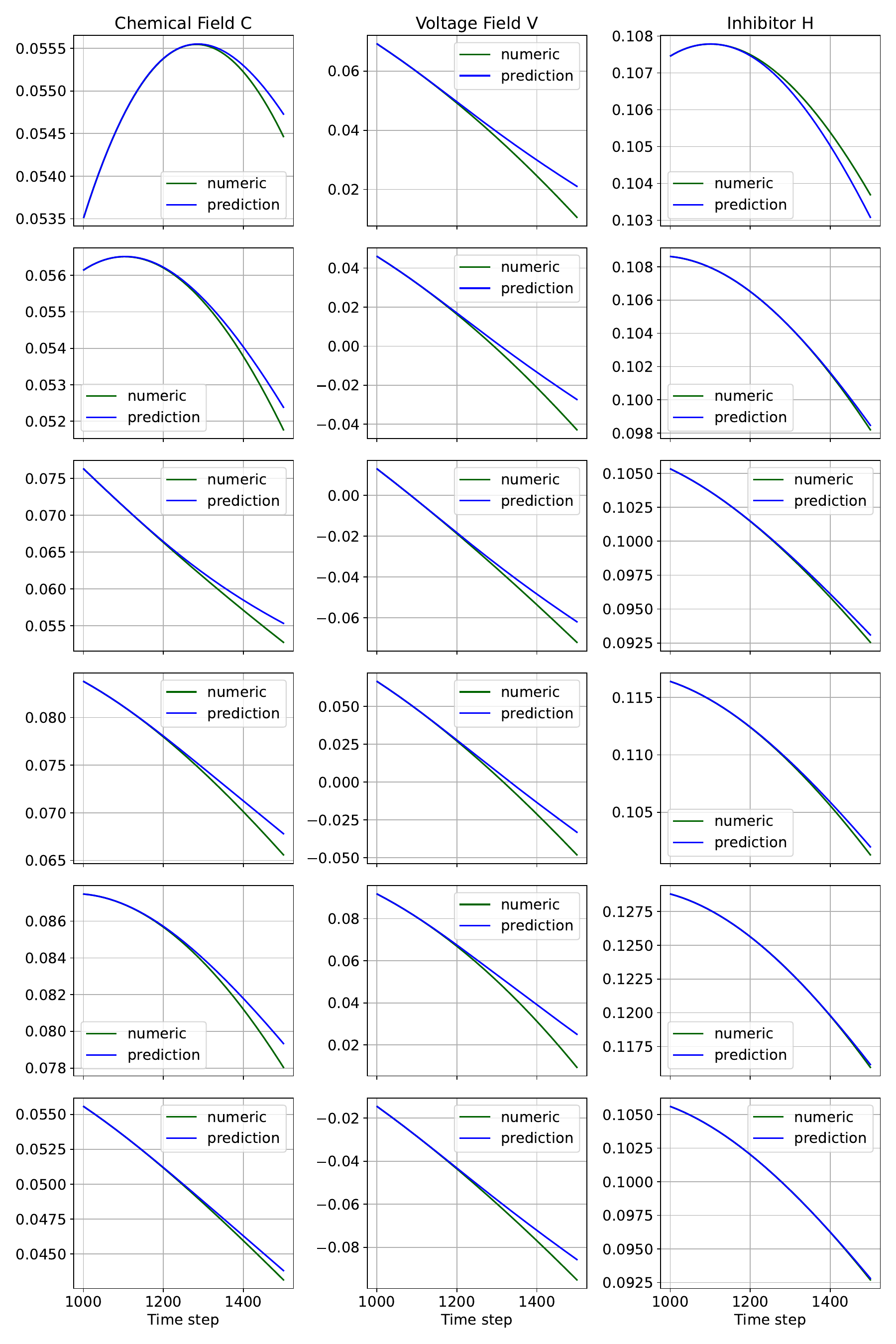}
\caption{\label{f:xtimecurves05}Numerical simulation versus ESN predictions from step 1000 to step 1500 and for 6 points on the cross-section $z=7$ for model version~3.}
\end{figure}

In Figure~\ref{f:xtimecurves}, similar curve comparisons are shown for model version~2, using the points depicted for that model version, compare Table~\ref{tb:comp-points}. Finally, the corresponding comparisons for model version~3 are presented in Figure~\ref{f:xtimecurves05}.

\clearpage
The Tables~\ref{tb:error-v1},~\ref{tb:error-v2} and~\ref{tb:error-v3} show the normalized root mean square error (NRMSE) calculated between the true numerical values of each model version and the corresponding ESN-predicted values at each model-relevant evaluation point and for the time step intervals shown in Figures~\ref{f:timecurves}, \ref{f:xtimecurves} and \ref{f:xtimecurves05}, respectively. Each row corresponds to the error at an individual point, the final row summarizes the total (global) NRMSE over all the model relevant evaluation points. 
\begin{table}[htbp]
  \begin{minipage}[t]{\linewidth}
  \centering
  \setlength{\tabcolsep}{7pt}
  \begin{tabular}{xxx}
  \hline\hline
  \multicolumn{3}{c}{NRMSE for predicted vs. true curves} \\
  \multicolumn{3}{c}{over 50 time steps, 4 Points as in Figure~\ref{f:timecurves}} \\
  C & V & H \\
  \hline
  5.69e-02 &   4.81e-02 &  7.47e-02 \\
  7.35e-02 &   4.31e-02 &  5.14e-02 \\
  5.62e-02 &   4.90e-02 &  3.37e-02 \\
  3.13e-02 &   3.43e-02 &  1.84e-01 \\
  \multicolumn{3}{c}{total NRMSE: 9.88e-03} \\
  \hline\hline
  \end{tabular}
  \end{minipage}
  \caption{\label{tb:error-v1}Normalized Root Mean Square Error (NRMSE) per point and the total error with respect to the curves of Figure~\ref{f:timecurves} for model version~1.}
\end{table}
\begin{table}[htbp]
  \begin{minipage}[t]{\linewidth}
  \centering
  \setlength{\tabcolsep}{7pt}
  \begin{tabular}{xxx}
  \hline\hline
  \multicolumn{3}{c}{NRMSE for predicted vs. true curves} \\
  \multicolumn{3}{c}{over 50 time steps, 6 Points as in Figure~\ref{f:xtimecurves}} \\
  C & V & H \\
  \hline
  4.29e-03 &  8.93e-03 & 5.34e-03 \\
  1.87e-03 &  7.24e-03 & 6.57e-03 \\
  4.57e-03 &  2.23e-03 & 6.60e-03 \\
  1.46e-02 &  5.32e-03 & 7.05e-03 \\
  1.22e-02 &  3.74e-03 & 8.29e-03 \\
  1.60e-02 &  1.56e-03 & 7.80e-03 \\
  \multicolumn{3}{c}{total NRMSE: 4.25e-04} \\
  \hline\hline
  \end{tabular}
  \end{minipage}
  \caption{\label{tb:error-v2}Normalized Root Mean Square Error (NRMSE) per point and the total error with respect to the curves of Figure~\ref{f:xtimecurves} for model version~2.}
\end{table}
\begin{table}[htbp]
  \begin{minipage}[t]{\linewidth}
  \centering
  \setlength{\tabcolsep}{7pt}
  \begin{tabular}{xxx}
  \hline\hline
  \multicolumn{3}{c}{NRMSE for predicted vs. true curves} \\
  \multicolumn{3}{c}{over 500 time steps, 6 Points as in Figure~\ref{f:xtimecurves05}} \\
  C & V & H \\
  \hline
  3.80e-02 & 8.19e-02 & 5.36e-02 \\
  5.08e-02 & 7.95e-02 & 7.38e-03 \\
  4.77e-02 & 5.11e-02 & 1.59e-02 \\
  5.24e-02 & 5.68e-02 & 1.67e-02 \\
  5.63e-02 & 8.75e-02 & 4.30e-03 \\
  2.17e-02 & 5.00e-02 & 2.19e-03 \\
  \multicolumn{3}{c}{total NRMSE: 1.32e-02} \\
  \hline\hline
  \end{tabular}
  \end{minipage}
  \caption{\label{tb:error-v3}Normalized Root Mean Square Error (NRMSE) per point and the total error with respect to the curves of Figure~\ref{f:xtimecurves05} for model version~3.}
\end{table}

The correspondence between numerical values and ESN predictions is generally good. Stronger divergences appear, for instance in Figure~\ref{f:timecurves}, around time step 150, coinciding with a period of heightened chaotic behavior. This is supported by the Lyapunov exponent graphic as indicated in Figure~\ref{f:lyapD}, which shows a local maximum at the same time point, suggesting increased sensitivity to initial conditions in the system’s dynamics.
Even tiny errors in the representation of the initial state can grow exponentially over time in chaotic systems. While ESNs are trained on observed patterns, the emergence of chaos may drive the system into regions of phase space that are underrepresented in the training data. As a result, prediction errors accumulate more rapidly, and the ESN’s output may diverge from the true trajectory, even if its earlier predictions were highly accurate. This behavior is not considered a flaw, but rather a well-known limitation of ESNs (and many data-driven models) when applied to deterministic chaotic systems.
A comparison of Figures~\ref{f:timecurves} and \ref{f:xtimecurves} reveals that the prediction horizon is longer for model version~2 (50 steps ahead) compared to approximately $35$ steps for model version~1. Moreover, the overall accuracy has improved, as indicated by the lower total error shown in Table~\ref{tb:error-v2} in comparison to~\ref{tb:error-v1}, respectively. Regarding model version~3, even though the accuracy is only slightly better than that of model~1 (see Table~\ref{tb:error-v3}), the prediction horizon is extended to 500 steps, which corresponds to a double prediction horizon than that of model 1 or 2  (compare Figure~\ref{f:xtimecurves05}). Furthermore, the predictions are accurate for nearly 250 steps corresponding to 50 steps in model 1 or 2. This comes at the cost of a double training time for model 3, that is, 1000 time steps which would be equivalent to 200 steps of training for model 1 or 2.
In the regions in which predictions for the first model version have been performed, the Lyapunov exponents are weakly positive ($< 0.001$). A low unpredictability can be normally expected. The system is weakly chaotic at best, it shows sensitivity to initial conditions, but this sensitivity grows slowly.
Systems with small positive Lyapunov exponents exhibit weak chaos, which permits somewhat longer, but still fundamentally limited, prediction horizons.
This is especially true when using noise-robust echo state networks, which can effectively model and generalize systems' dynamics. Being able to predict 30 steps ahead in a weakly chaotic regime is generally considered a good level of efficiency, especially for complex systems like those in electrophysiology, involving coupled PDEs, nonlinear feedback, and multiple interacting variables like the fields $c$, $v$ and $h$.
Model version~2 enables trustworthy predictions up to $50$ steps into the future, while the Lyapunov exponents raise well above $0.001$ in the interval of calculations. For model version~3, predictions are trustworthy over almost 350 to 500 steps ahead in the future. This result is particularly notable given that the Lyapunov exponents, especially those associated with the chemical field, remain consistently and significantly above $0.001$.
\section{Conclusions}
The performance of the ESN, in all its three versions is particularly notable and demonstrates the ability to generate reliable predictions several time steps into the future (see Figures~\ref{f:timecurves}, \ref{f:xtimecurves}, \ref{f:xtimecurves05}).
The prediction horizons of approximately 30 to 40 steps of $\Delta t=0.01$ for model version 1, 50 steps of $dt=0.01$ for version 2, and up to 500 steps of $\Delta t=0.002$ for version 3, indicate a substantial efficiency improvement.
Long-term predictive capability is especially valuable when timely control inputs are required to steer a complex system toward desired target states.
Importantly, the predictions presented here do not concern a one-dimensional time series or a small number of action potentials, but rather the simultaneous prediction of $30\times 30\times 30\times 3=$ 81,000 field components, corresponding to three fields discretized over the full three-dimensional calculation domain.
Each of these fields can be interpreted as a one-dimensional time series with 27,000 spatial values per time step, or equivalently, as an individual function evolving over time.
The ESN thus carries out a large-scale, parallel prediction task over a high-dimensional spatiotemporal system.
Recent studies have successfully employed Echo State Networks for long-term prediction of electrophysiological signals. For instance, Shahi et al.~\cite{b17} achieved accurate predictions of 20 action potentials using an architecture that integrated a long short-term memory (LSTM) autoencoder into an ESN framework.
Cardiac action potential (AP) time series are, in most experimental and modeling contexts, measured at a single point (or cell) in the heart and are represented as one-dimensional (1D) time series.
Prediction of 20 action potentials ahead, at a sampling rate of 10~kHz, translates to approximately 70,000 prediction steps into the future. This represents a remarkably long prediction horizon for a physiological time series, particularly given the nonlinear and noisy nature of cardiac signals. The model’s ability to maintain accuracy over such an extended sequence highlights the strength of the autoencoder-ESN hybrid architecture.

The task addressed in this work is quite complex. Specifically, a high-dimensional spatiotemporal system governed by a coupled set of differential equations with 81,000 individual field components modeling electrophysiological dynamics, has been considered.
The interactions of these components are governed by nonlinear, partly weak chaotic dynamics which represents a significant scaling of the prediction task, from forecasting a single signal to simultaneously predicting many thousands of spatially distributed, interdependent signals across time. The 81k-field model, discussed here, handles a vast coupled system with a minimal ESN architecture and can be considered a remarkable achievement. 3D electrophysiology models are widely used in cardiac research and clinical applications. They should be able to capture spatial heterogeneity of cardiac tissue, simulate wavefront propagation, reentry, and block key mechanisms in arrhythmias.
They should also enable non-invasive personalization using imaging and ECG data and provide a testbed for therapies like pacing, ablation, and defibrillation. Therefore, these models deserve attention and further development~\cite{b19}.

\paragraph{Future Work:}
The use of ESNs in biological modeling, particularly in gene regulatory networks, signaling cascades, and protein dynamics, remains limited compared to other machine learning approaches, such as LSTM or mechanistic ODE-based models. ESNs excel at capturing temporal dependencies, feedback loops, and nonlinear dynamics, which are central to morphogenesis and gene regulation. Because ESNs are model-free by design, they don’t inherently encode physical laws or spatial structure which limits their direct applicability to systems governed by PDEs, reaction-diffusion models, or mechanical constraints.
Recent work explores physics-informed ESNs, spatially structured reservoirs, and hybrid models that combine ESNs with mechanistic frameworks. Approaches in this direction are to be followed in our future work~\cite{b20}.
The model presented here, being a low- to medium-resolution configuration, is more suitable for proof-of-concept investigations. For realistic applications, it would need to be improved and extended, for example by increasing resolution or model complexity.
Input preprocessing (such as scaling) and a systematic optimization, for instance, via evolutionary algorithms or gradient-free methods, belong to the immediate development plans.
Also, adding output-to-reservoir feedback, or feedback connections for richer dynamics, should be a necessary further development step.
To reliably simulate morphogenesis in a 3D model, discretization must ideally reach cellular resolution, which typically involves thousands to millions of elements depending on the biological scale and complexity. This necessitates larger models and requires parallel or distributed implementations to handle high-resolution simulations efficiently. Hierarchical or coarse-to-fine modeling strategies may be needed to balance computational cost with biological realism. Equally important is the integration of domain knowledge into the reservoir design. Sensitivity analyses to identify critical parameters, as well as cross-validation with experimental data, are also planned as part of our model further development.

\paragraph{System Configuration and Runtimes:}
All models are programmed and the results are visualized in Python (version 3.13) as Jupyter notebooks, using the Numpy (version 2.2), ReservoirPy (version 0.4.1), and Matplotlib (version 3.10) libraries.
The computer used a recent eight core AMD Zen CPU with 64 GByte memory.
The runtimes of the models were below 30 seconds.


\begin{thebibliography}{99}\clubpenalty10000\widowpenalty10000
%
\providecommand{\bibTitle}[1]{\emph{#1}}
\providecommand{\asDOI}[1]{doi:\href{https://doi.org/#1}{#1}}
%
\bibitem{b11}
   Ambekar, Z.:
   \bibTitle{Transforming Morphogenesis: Using Deep Learning to Understand Biological Development},
   Medium weblog,
  \url{https://medium.com/@zeneil_writes/transforming-morphogenesis-using-deep-learning-to-understand-biological-development-e4af7b6b10fc},
  2025.
  %
\bibitem{b10}
   Avilés, E.E.O et al.:
  \bibTitle{Neural network Approximations for Reaction\hyp{}Diffusion Equations -- Homogeneous Neumann Boundary Conditions and Long-time Integrations},
  arXiv preprint,
  \asDOI{10.48550/arXiv.2409.08941},
  2024.
  %
\bibitem{b09}
  Chen, S. et al.:
  \bibTitle{Learning Interactions in Reaction Diffusion Equations by Neural Networks},
  Entropy 25(3), 489,
  \asDOI{10.3390/e25030489}
  2023.
  %
\bibitem{b16}
  Doan, N.A.K. et al.:
  \bibTitle{Physics-Informed Echo State Networks},
  J Comput Science 47, 101237,
  \asDOI{10.1016/j.jocs.2020.101237},
  2020.
  %
\bibitem{b14}
  Duggento, A.  et al.:
  \bibTitle{Recurrent neural networks for reconstructing complex directed brain connectivity},
  41st Annual International Conference of the IEEE Engineering in Medicine and Biology Society (EMBC),
  \asDOI{10.1109/EMBC.2019.8856721},
  2019.
  %
\bibitem{b13}
  Duggento, A.  et al.:
  \bibTitle{Echo state network models for nonlinear Granger causality},
  Philosophical Transactions of the Royal Society A,
  \asDOI{10.1098/rsta.2020.0256},
  2021.
  %
\bibitem{b003}
   FitzHugh, R.:
  \bibTitle{Impulses and Physiological States in Theoretical Models of Nerve Membrane},
   Biophys J 1(6), pp 445-–466,
   \asDOI{10.1016/s0006-3495(61)86902-6},
   1961.
   %
\bibitem{b12}
  Ibanez-Soria, D. et al.:
  \bibTitle{Detection of Generalized Synchronization using Echo State Networks},
  arXiv preprint,
  \asDOI{10.48550/arXiv.1710.08286},
  2017.
  %
\bibitem{b05}
  Lagergren, J.H. et al.:
  \bibTitle{Biologically-informed neural networks guide mechanistic modeling from sparse experimental data},
  PLoS Comput Biol 16(12):e1008462,
  \asDOI{10.1371/journal.pcbi.1008462},
  2020.
  %
\bibitem{b07}
  Lu, L. et al.:
  \bibTitle{A deep learning library for solving differential equations},
  SIAM Review 63(1), pp 208--228,
  \asDOI{10.1137/19M1274067},
  2021.
  %
\bibitem{b06}
  Mandal, S. and Aihara, K.:
  \bibTitle{Revisiting multifunctionality in reservoir computing},
  arXiv preprint,
  \asDOI{10.48550/arXiv.2504.12621},
  2025.
  %
\bibitem{b04}
  Mircea, M. et al.:
  \bibTitle{Inference of dynamical gene regulatory networks from single-cell data with physics informed neural networks},
  arXiv preprint,
  \asDOI{10.48550/arXiv.2401.07379},
  2021.
  %
\bibitem{b004}
  Nagumo, J. et al.:
  \bibTitle{An Active Pulse Transmission Line Simulating Nerve Axon},
  in: Proceedings of the IRE 50(10), pp 2061--2070,
  \asDOI{10.1109/JRPROC.1962.288235},
  1962.
  %
\bibitem{b01}
  Ren, Z. et al.:
  \bibTitle{Physics-Informed Neural Networks: A Review of Methodological Evolution, Theoretical Foundations, and Interdisciplinary Frontiers Toward Next-Generation Scientific Computing},
  Appl Sci 15(14), 8092,
  \asDOI{10.3390/app15148092},
  2025.
  %
\bibitem{b17}
  Shahi, S. et al.:
  \bibTitle{A Machine-learning Approach for Long-term Prediction of Experimental
  Cardiac Action Potential Time Series Using an Autoencoder and Echo
  State Networks},
  Chaos 32(6):063117,
  \asDOI{10.1063/5.0087812},
  2022.
  %
\bibitem{b03}
  Smith, C.A. and Yates, C.A.:
  \bibTitle{Incorporating domain growth into hybrid methods for reaction-diffusion systems},
  J R Soc Interface 18(177):20201047,
  \asDOI{10.1098/rsif.2020.1047},
  2021.
  %
\bibitem{b18}
  Sun, C. et al.:
  \bibTitle{A Review of Designs and Applications of Echo State Networks},
  arXiv preprint,
  \asDOI{10.48550/arXiv.2012.02974},
  2020.
  %
\bibitem{b23}
  Trouvain, N. et al.:
  \bibTitle{\emph{ReservoirPy}: An Efficient and User-Friendly Library to Design Echo State Networks},
   in: Artificial Neural Networks and Machine Learning -- ICANN 2020, Lect Notes Comput Sci, vol 12397,
   \asDOI{10.1007/978-3-030-61616-8\_40},
   2020.
   %
\bibitem{b02}
  Turing, A. M.
  \bibTitle{The Chemical Basis of Morphogenesis},
  Philos Trans R Soc Lond B Biol Sci, 237(641), pp 37--72,
  \asDOI{10.1098/rstb.1952.0012},
  1952.
  %
\bibitem{b20}
  Werneck, Y.B. et al.:
  \bibTitle{Replacing the FitzHugh-Nagumo Electrophysiology Model by Physics-Informed Neural Networks},
  in: Computational Science -- ICCS 2023, Lect Notes Comput Sci, vol 14074, 
  \asDOI{10.1007/978-3-031-36021-3\_67},
  2023.
  %
\bibitem{b15}
  Yang, L.  et al.:
  \bibTitle{Brain-inspired modular echo state network for EEG-based emotion recognition},
  Front Neurosci 18:1305284,
  \asDOI{10.3389/fnins.2024.1305284},
  2024.
  %
\bibitem{b19}
  Ye, Y. et al.:
  \bibTitle{A Spatial-Temporally Adaptive PINN Framework for 3D Bi-Ventricular Electrophysiological Simulations and Parameter Inference},
  in: Medical Image Computing and Computer Assisted Intervention -- MICCAI 2023, Lect Notes Comput Sci, vol 14226,
  \asDOI{10.1007/978-3-031-43990-2\_16},
  2023.
  %
\bibitem{b08}
  Zubov, K. et al.:
  \bibTitle{NeuralPDE: Automating Physics-Informed Neural Networks (PINNs) with Error Approximations},
  arXiv preprint,
  \asDOI{10.48550/arXiv.2107.09443},
  2021.
  %
\end{thebibliography}
\end{document}